\PassOptionsToPackage{table}{xcolor}
\documentclass[10pt,twocolumn,letterpaper]{article}

\usepackage{iccv}
\usepackage{times}
\usepackage{epsfig}
\usepackage{graphicx}
\usepackage{amsmath}
\usepackage{amssymb}

\usepackage{booktabs}
\usepackage{multirow}
\usepackage[export]{adjustbox}
\usepackage{threeparttable}
\usepackage{float}
\usepackage{subcaption}
\usepackage{tabularx}
\usepackage{adjustbox}
\usepackage{color}
\usepackage{graphicx, amsmath, amssymb, multirow, overpic, textpos}
\usepackage[table]{xcolor}

\renewcommand{\paragraph}[1]{\vspace{1.25mm}\noindent\textbf{#1}}

\newlength\savewidth
\definecolor{baselinecolor}{gray}{.9}

\newcommand{\hatm}{\hat{m}}

\newcommand{\hy}{\hat{y}}

\newcommand{\hp}{\hat{p}}

\usepackage[misc]{ifsym} 

\definecolor{linkcolor}{RGB}{255,0,0}
\definecolor{urlcolor}{RGB}{255,105,180}
\definecolor{citecolor}{RGB}{66,168,235}
\usepackage[pagebackref,breaklinks,colorlinks]{hyperref}
\hypersetup{colorlinks=true,linkcolor=linkcolor,urlcolor=urlcolor,citecolor=citecolor}

\usepackage[capitalize]{cleveref}
\crefname{section}{Sec.}{Secs.}
\Crefname{section}{Section}{Sections}
\Crefname{table}{Table}{Tables}
\crefname{table}{Tab.}{Tabs.}

\iccvfinalcopy 

\ificcvfinal\fi

\begin{document}


\title{Tube-Link: A Flexible Cross Tube Framework for Universal Video Segmentation}
\author{
Xiangtai Li$^{1}$ \quad
Haobo Yuan$^{1}$ \quad
Wenwei Zhang$^{1,4}$ \quad \\
Guangliang Cheng$^{2,3}$ \quad
Jiangmiao Pang$^{4}$ \quad
Chen Change Loy$^{1 \textrm{\Letter}}$ 
\\[0.1cm]
\small $ ^1$ S-Lab, Nanyang Technological University \quad 
\small $ ^2$ University of Liverpool \quad $ ^3$ SenseTime Research \quad   \small $ ^4$ Shanghai AI Lab \\
{\tt\small \{xiangtai.li, ccloy\}@ntu.edu.sg} \\
 \url{https://github.com/lxtGH/Tube-Link}
}

\maketitle

\ificcvfinal\thispagestyle{empty}\fi

\begin{abstract}
Video segmentation aims to segment and track every pixel in diverse scenarios accurately. 
In this paper, we present Tube-Link, a versatile framework that addresses multiple core tasks of video segmentation with a unified architecture. 
Our framework is a near-online approach that takes a short subclip as input and outputs the corresponding spatial-temporal tube masks. 
To enhance the modeling of cross-tube relationships, we propose an effective way to perform tube-level linking via attention along the queries. 
In addition, we introduce temporal contrastive learning to instance-wise discriminative features for tube-level association.
Our approach offers flexibility and efficiency for both short and long video inputs, as the length of each subclip can be varied according to the needs of datasets or scenarios. 
Tube-Link outperforms existing specialized architectures by a significant margin on five video segmentation datasets. 
Specifically, it achieves almost 13\% relative improvements on VIPSeg and 4\% improvements on KITTI-STEP over the strong baseline Video K-Net. When using a ResNet50 backbone on Youtube-VIS-2019 and 2021, Tube-Link boosts IDOL by 3\% and 4\%, respectively. Code is available at \url{https://github.com/lxtGH/Tube-Link}.
\end{abstract}
\section{Introduction}


\begin{figure}[t!]
	\centering
	\includegraphics[width=1.0\linewidth]{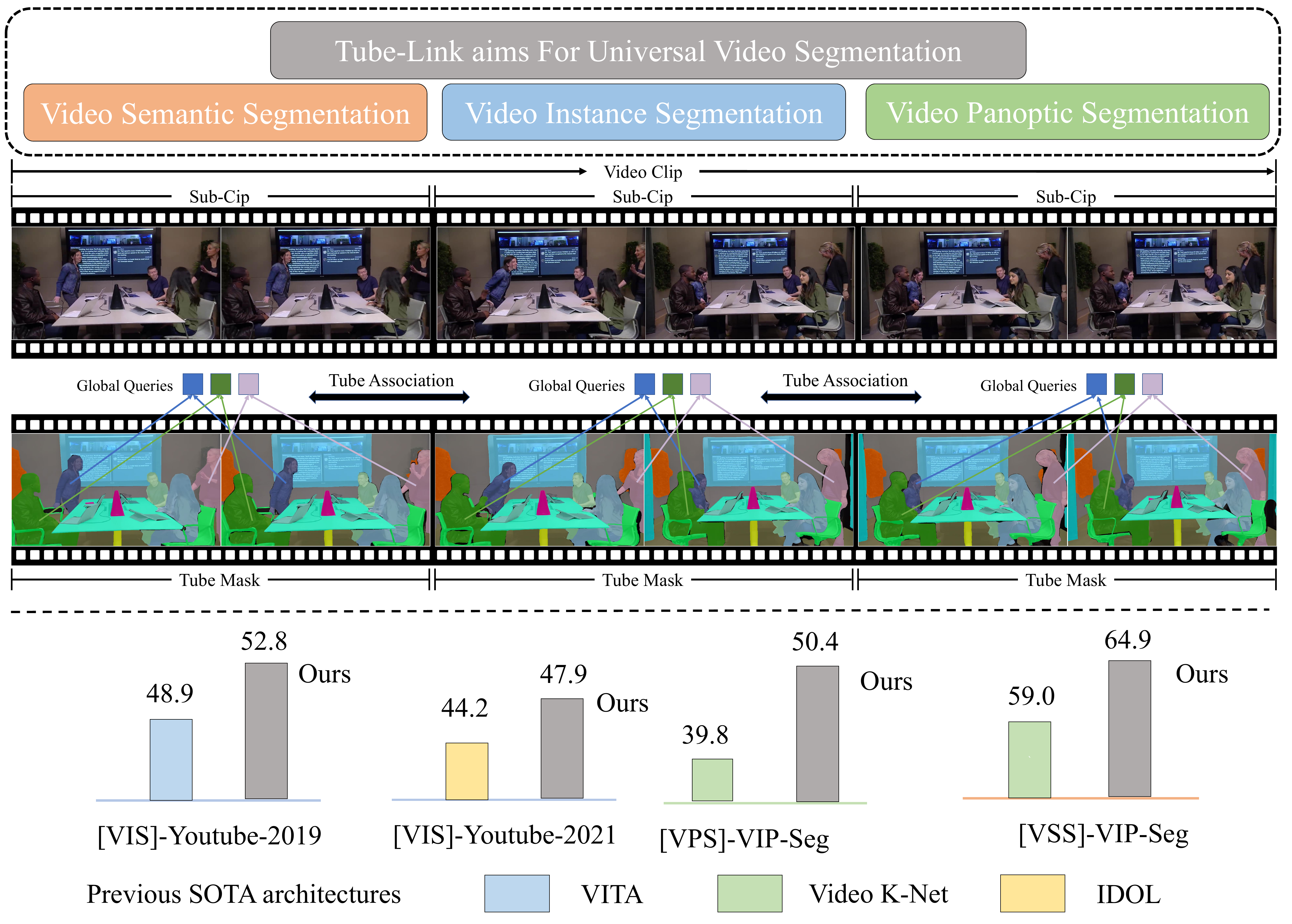}
	\caption{\small Tube-Link takes subclips as inputs and links the resulting tubes in a near online manner. Our design embraces flexibility, efficiency, and temporal consistency, making it suitable for various video segmentation tasks, including VSS, VIS, and VPS. Notably, our method outperforms the best-specialized architectures for these tasks on multiple datasets. Best viewed in color.}
    \label{fig:teasear}
\end{figure}

%
The success of the Detection Transformer (DETR)~\cite{detr} has inspired recent works~\cite{cheng2021mask2former,zhang2021knet,cheng2021maskformer,wang2020maxDeeplab} to develop universal architectures for addressing all image segmentation tasks using the same architecture, also known as universal image segmentation. 
In video segmentation, the Video Panoptic Segmentation (VPS) task involves segmenting and tracking each pixel in input video clips~\cite{kim2020vps,STEP,hurtado2020mopt}, unifying the Video Semantic Segmentation (VSS)~\cite{miao2021vspw} and Video Instance Segmentation (VIS)~\cite{vis_dataset} tasks. 
To minimize specialized architecture design for each task, recent studies~\cite{li2022videoknet,kim2022tubeformer} follow a universal approach to video segmentation by solving sub-tasks of VPS in a single unified framework. 
These methods typically use an end-to-end set prediction objective and successfully address multiple tasks without modifying the architecture and loss.

While recent studies have demonstrated promising results, there are still several issues with VPS models and universal video segmentation methods. 
One major challenge is the lack of exploration of VPS for arbitrary scenes and video clip lengths. To address this gap, Miao~\etal~\cite{miao2022large} have introduced a more challenging benchmark, named VIPSeg, which features long videos and diverse indoor and outdoor scenes. This new dataset presents new challenges to existing VPS methods~\cite{kim2020vps,miao2022large,li2022videoknet,ViPDeepLab}, such as increased occlusions and appearance changes in diverse scenarios. 
Another issue with universal methods~\cite{li2022videoknet,kim2022tubeformer} is that they cannot achieve comparable results to recent Transformer-based VIS methods~\cite{IDOL,seqformer}, which raises the question of whether we can design a universal video segmentation method to avoid these specialized designs.

To gain a better understanding of the limitations of current solutions for video segmentation, we examine the existing methods and categorize them into two groups based on how they process input video clips: \textit{online} and \textit{near-online}. The former~\cite{kim2020vps,li2022videoknet,IDOL,vis_dataset,woo2021learning_associate_vps} performs video segmentation at the frame level, while the latter~\cite{ViPDeepLab,STEP,kim2022tubeformer,VIS_TR,cheng2021mask2former,hwang2021video,seqformer} processes clip-wise inputs and directly obtains tube-wise masks. 
However, there are trade-offs in deploying either approach. Although online methods offer great flexibility, they struggle to use temporal information effectively and thus compromise segmentation performance. 
On the other hand, near-online methods achieve better segmentation quality, but they cannot handle long video clips, and most approaches are only validated in VIS tasks, which have fewer instances and simpler scenes.


In this study, we introduce Tube-Link, a universal video segmentation framework that combines the benefits of both online and near-online methods. 
The framework follows a common input and output space for video segmentation tasks, where a long clip input is split into multiple subclips. 
Each subclip contains several frames within a temporal window, and the output is a spatial-temporal mask that tracks the target entity. 
Our framework is compatible with contemporary methods such as Mask2Former-VIS~\cite{cheng2021mask2former_vis}, where each global query encodes the same tracked entity, and the global queries perform cross-attention with spatial-temporal features in the decoder directly.

In particular, we propose several key improvements to the Mask2Former-VIS meta-architecture. \textbf{First}, we extend the instance query into an entity query (either thing or stuff) to improve temporal consistency for both thing and stuff segmentation, thus generalizing Mask2Former-VIS into a universal segmentation architecture. 
\textbf{Second}, we improve the modeling of cross-tube relationships from two different aspects through temporal consistency learning and temporal association learning. 
For the former, we design a simple link head with self-attention layers that links global queries across tubes to enforce segmentation consistency across tubes. 
For the latter, we generalize previous frame-level contrastive learning into tube-level and learn temporal association embeddings with a temporal contrastive loss. 
Unlike previous works~\cite{li2022videoknet,IDOL,qdtrack} that only learn from two adjacent frames, we consider multiple frames to learn cross-tube consistency. The learned embeddings are then used to perform tube mask matching, which is much more effective than previous counterparts~\cite{li2022videoknet} in complex video scenarios. 
\textbf{Third}, with the flexibility of window size and learned association embeddings, we show that one can enlarge the subclip size to improve temporal consistency and inference efficiency, even when trained with fewer subclip inputs. 

Our approach is a simple yet flexible framework that outperforms specialized architectures across various video segmentation tasks.
We evaluate Tube-Link on three video segmentation tasks using six datasets (VIP-Seg~
\cite{miao2022large}, KITTI-STEP~\cite{STEP}, VSPW~\cite{miao2021vspw}, YouTube-VIS-2019/2021~\cite{vis_dataset}, OVIS~\cite{qi2022occluded}). We demonstrate that, for the first time, our single architecture performs on par or better than the most specialized architectures on five video benchmarks. In particular, as shown in Fig.~\ref{fig:teasear}, using the same ResNet-50 backbone, Tube-Link outperforms recently published works Video K-Net~\cite{li2022videoknet} 4\% VPQ on KITTI-STEP, 13\% VPQ, and 8\% STQ on VIP-Seg, VITA~\cite{heo2022vita} and IDOL~\cite{IDOL} on YouTube-VIS-2019 by 3\% mAP. We also outperform TubeFormer~\cite{kim2022tubeformer} on VPSW by 1.7\% mIoU, and on YouTube-VIS-2019 by 5.3\% mAP. 


\section{Related Work}
\label{sec:related_work}

\noindent
\textbf{Specialized Video Segmentation.} VSS aims to predict a class label for each pixel in a video. 
Recent approaches~\cite{shelhamer2016clockwork,DFF,miao2021vspw,sun2022vss,sun2022mining} model the temporal consistency or acceleration using methods such as optical flow warping or spatial-temporal attention. 
VIS~\cite{vis_dataset} extends instance segmentation into video, aiming to segment and track each object simultaneously. Several methods~\cite{mask_pro_vis,lin2021video, zhu2022instance, qin2022graph,li2021improving} link instance-wise features in the video. 
Recent studies~\cite{VIS_TR,hwang2021video,seqformer,IDOL,TeViT,MeViS} have extended DETR into VIS, proposing better ways to fuse different queries along the temporal dimension. 
However, these methods cannot be directly transferred to complex scenes~\cite{STEP,miao2022large} due to the limited instances and simpler scenes used in their training. 
VPS aims to generate instance tracking IDs and panoptic segmentation results across video clips. Kim~\etal~\cite{kim2020vps} mainly focus on short-term tracks, using only six frames for each clip in Cityscapes video sequences. 
STEP~\cite{STEP} proposes a Segmentation and Tracking Quality (STQ) metric that decouples the segmentation and tracking error, along with long sequence VPS datasets. 
Recently, VIP-Seg~\cite{miao2022large} proposed a more challenging dataset containing various scenes, scales, instances, and clip lengths. However, current solutions~\cite{kim2020vps,woo2021learning_associate_vps,li2022videoknet} for VPS mainly focus on online or near-online approaches, which have difficulty adapting to general tasks (VSS, VIS) or handling the complex scenarios in VIP-Seg dataset.

\noindent
\textbf{Universal Architectures For Segmentation.} 
Recent studies~\cite{wang2020maxDeeplab,zhang2021knet,cheng2021mask2former,yu2022kmaxdeeplab,panopticpartformer,yuan2021polyphonicformer} adopt mask classification architectures with an end-to-end set prediction objective for universal segmentation, achieving better results than specialized models. 
In video segmentation, two representative works are Video K-Net~\cite{li2022videoknet} and TubeFormer~\cite{kim2022tubeformer}. The former unifies the video segmentation pipeline via kernel tracking and linking, while the latter adopts a near-online approach and obtains tube-level masks with cross-attention along the tube features and queries. 
However, both works fail to replace specialized models, as their performance on specific tasks or datasets is still worse than the best-specialized architecture (they perform worse in VIS). 
Our proposed Tube-Link is the first generalized architecture that outperforms existing state-of-the-art specialized architectures on three video segmentation tasks. 
In comparison to~\cite{kim2022tubeformer}, Tube-Link further explores cross-tube association, which is critical in long and complex scenes.

\noindent
\textbf{Video Object Detection and Tracking.}
Object tracking is a crucial task in VPS, and many works adopt the tracking-by-detection paradigm~\cite{bewley2016simple,leal2016learning,xu2019spatial,zhu2018online,porzi2020learning}. 
These methods divide the task into two sub-tasks, where objects are first detected by an object detector and then associated using a tracking algorithm. 
Recent works~\cite{park2022per,MOSE} also perform clip-wise segmentation or tracking~\cite{cliptrack}. However, the former only focuses on single-object mask tracking, while the latter considers global tracking via clip-level matching. 
Our proposed Tube-Link naturally handles both settings yet provides additional segmentation masks as outputs.
In Video Object Detection, the literature~\cite{zhu17fgfa,deng19rdn,chen18stlattice,xiao18stmn,zhou2022transvod} has seen the broad usage of information across multiple frames. 
Our work is related to TransVOD~\cite{zhou2022transvod}, which uses a local temporal window. However, in these studies, learning correspondences and links across tubes are not explored, partly due to the nature of ImageNet-VID dataset~\cite{russakovsky2015imagenet}, which does not require object tracking and segmentation.

\section{Methodology}

\subsection{Preliminary and Motivation}

In this subsection, we begin by introducing a unified notation for universal video segmentation (VSS, VIS, and VPS), and then demonstrate how this task can be modeled through linking the tracked short tube masks. 
After that, we further motivate tube-wise matching by comparing it against conventional frame-wise matching. 

\noindent
\textbf{Universal Video Segmentation Formulation.} We denote a video clip input as $ V \in \mathbb{R}^{T\times H\times {W}\times 3}$, where $T$ represents the frame number and ${H}\times {W}$ are the spatial size.
The video clip is annotated with segmentation masks. The masks of a particular entity can be linked along the time dimension to form a tube. 
The annotations are denoted as $\{y_i\}_{i=1}^G = \{(m_i, c_i)\}_{i=1}^G \,$, where the $G$ is the number of ground truth, each tube mask $m_i \in {\{0,1\}}^{{T}\times {H}\times {W}}$ does not overlap with each other, and $c_i$ denotes the ground truth class label of the i-th tube. The background is assigned with value 0, and the foreground masks are assigned to be 1 in each tube mask $m_i$. 
VPS requires temporally consistent segmentation and tracking results for each pixel. 
Specifically, a model makes predictions on a set of video clips $\{\hy_i\}_{i=1}^N = \{(\hatm_i, \hp_i(c))\}_{i=1}^N$, where $\hatm_i \in {[0,1]}^{T\times H\times W}$ denotes the predicted tube, and $\hp_i(c)$ denotes the probability of assigning class $c$ to a clip $\hatm_i$ belonging to a predefined category in a set $C$. 
The number of entities is given by $N$, which includes countable thing classes and countless stuff classes. 
In particular, $i$ is the tracking ID for the thing class. 
When $N=C$ and $C$ only contain stuff classes, VPS turns into VSS. 
If ${\{\hy_i\}_{i=1}^N}$ can overlap and $C$ only contains the thing classes, VPS turns into VIS. 
We use such notations for universal video segmentation.

\noindent
\textbf{Video Segmentation as Linking Short Tubes.} 
To segment a video clip $V$, we divide it into a set of $L$ smaller subclips: ${\{v_{i}\}}_{i=1}^N$, where $v_{i} \in \mathbb{R}^{n\times H\times {W}\times 3}$, and $ n = T / L$ with $n$ representing the window size along the temporal dimension. 
The window size is flexible and can be adjusted according to the dataset. 
By taking subclips as inputs, we obtain shorter segmentation tubes than the whole clip. We perform tracking and linking across nearby tubes. Each predicted tube is represented as $\{\hat{y}_{i}^{t}\}_{i=1}^N = \{(\hatm_i^{t}, \hp_i(c)^{t})\}_{i=1}^N$, where $t$ is the index of each small tube, $\hatm_i^{t} \in {[0,1]}^{n\times H\times W}$ represents the segmentation masks, and $\hp_i(c)^{t}$ is the corresponding category. 
The final video prediction is obtained by linking each tube, $\{\hy_i\}_{i=1}^N = {\mathrm{Link}}({\{\hat{y}_{i}^{t}\}_{i=1}^N})_{t=1,..L}$. 
In our formulation, tracking is only performed across different tubes, and the segmentation within each tube is assumed to be consistent. Note that the key to achieving temporally consistent segmentation is the design of function $\rm{Link}$.

\begin{table}[!t]
	\centering
	\caption{\small \textbf{Exploration experiment on tube-wise matching.} Youtube-VIS: mAP. VIP-Seg:VPQ. We directly use pre-trained models by changing the input to two consecutive frames. }
	\label{tab:toy_exp}
  \scalebox{0.62}{
    \begin{tabular}{ r c c c}
    \toprule[0.15em]
     Method & Youtube-VIS-2019 & Youtub-VIS-2021 & VIP-Seg \\
    \toprule[0.15em]
    Min-VIS~\cite{huang2022minvis} & 47.4 &  44.2 & -\\
    Min-VIS + tube matching & 48.8 (+1.4) & 45.5 (+1.3)& - \\
    Video K-Net~\cite{li2022videoknet} & - & - & 26.1 \\
    Video K-Net + tube matching & - & - & 27.6 (+1.5) \\
    \bottomrule[0.2em]
    \end{tabular}
}
\end{table}

\noindent
\textbf{Motivation of Tube-wise Matching.} 
Existing video segmentation methods~\cite{IDOL,huang2022minvis,li2022videoknet} often perform instance association via frame-wise matching. This approach ignores local temporal information and can lead to occlusion errors. In this study, we propose to perform tube-wise matching using global queries based on their corresponding trackers. To motivate our approach, we modify the input of two representative works, Min-VIS~\cite{huang2022minvis} and Video K-Net~\cite{li2022videoknet}, by replacing the single frame input with a subclip input. Each subclip contains two frames. Without additional re-training or computation costs, we observe consistent improvements on three video segmentation datasets and two video segmentation tasks, VIS and VPS, as shown in Table~\ref{tab:toy_exp}. These findings suggest that cross-tube information is worth exploring for achieving universal segmentation. Therefore, we propose to model the $\rm{Link}$ function as tube-wise matching and linking, and term our framework Tube-Link.

\begin{figure}[t!]
	\centering
	\includegraphics[width=1.0\linewidth]{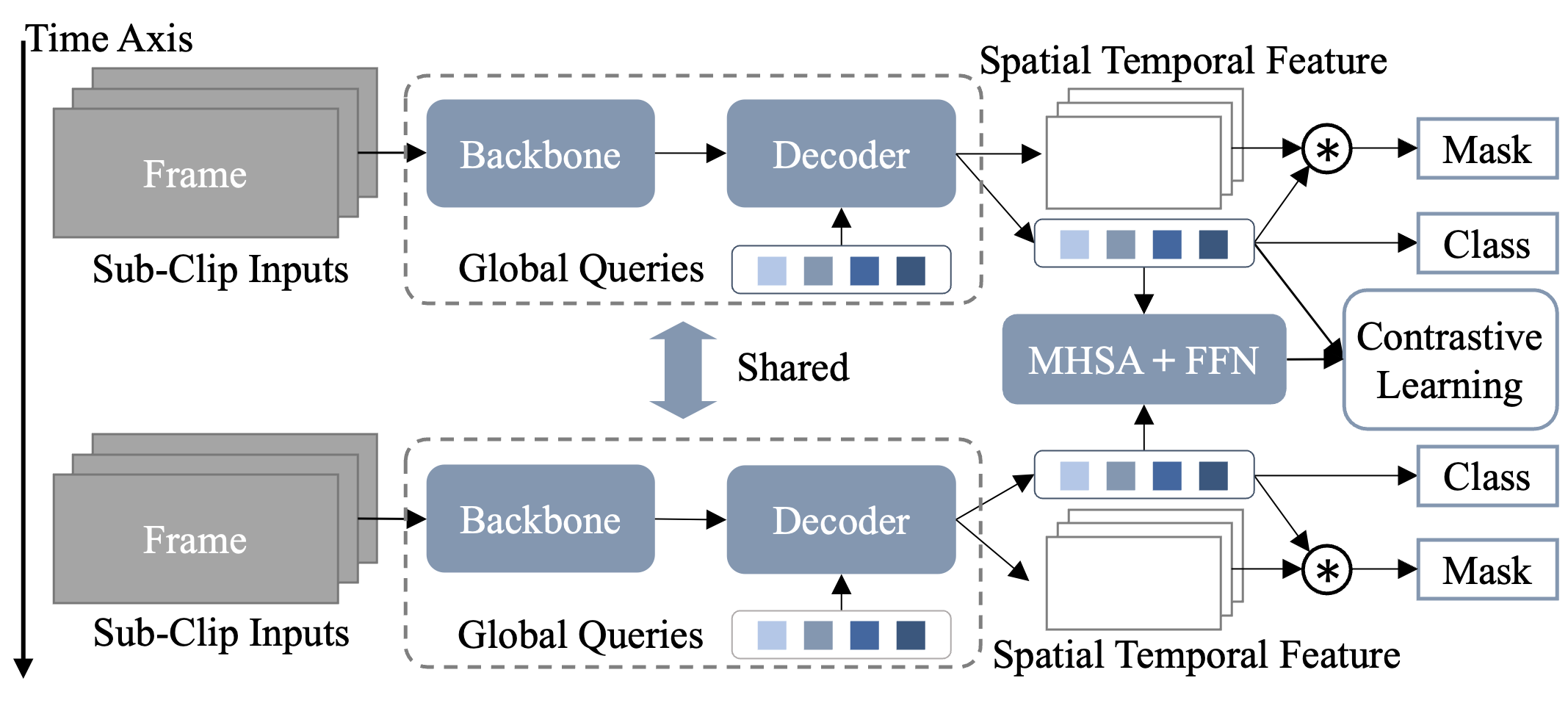}
	\caption{\small Our proposed Tube-Link framework. Given the subclips as input, we first use global queries to perform within-tube spatial-temporal learning to obtain the tube masks and labels. Then we link the object queries with cross-tube contrastive learning and self-attention.} 
	\label{fig:method}
    \vspace{-5mm}
\end{figure}

\subsection{Tube-Link Framework}
\label{sec:tb_framework}

We first introduce our extension to Mask2Former-VIS, which will serve as a baseline.
Then, we present two improvements, including cross-tube Temporal Contrastive Learning (TCL) and Cross-Tube Linking (CTL), to better model cross-tube relations. 
The training process of our method is illustrated in Fig.~\ref{fig:method}.

\noindent
\textbf{Mask2Former-VIS Extension as Baseline.} Following~\cite{VIS_TR,cheng2021mask2former_vis}, given a subclip $v_{i}$, we first employ Mask2Former-VIS~\cite{cheng2021mask2former_vis} to extract the spatial-temporal feature $\mathbf{F}_{i} \in \mathbb{R}^{n \times C\times H\times {W}}$. 
%
Mask2Former uses multiscale features and a cascaded decoder to perform cross-attention. Thus, we denote the layer index $l$ to indicate the layer number.
The \textit{global queries}, $\mathbf{Q}_{l-1} \in \mathbb{R}^{N_{q} \times C}$, perform masked cross-attention between $\mathbf{F}_{i}$ and $\mathbf{Q}_{i}$ as follows,
\begin{align}
    \label{equ:sp_attention}
    \mathbf{Q}_{l} = \mathrm{softmax}(\mathbf{M}_{l-1} + \mathrm{MLP}(\mathbf{Q}_{l-1})\mathbf{K}_{l}^{T})\mathbf{V}_{l} + \mathbf{Q}_{l-1},
\end{align}
where $N_{q}$ is the query number and is set to 100 by default, $\mathbf{M}_{l-1} \in \mathbb{R}^{n \times H\times {W}} $ is the binarized output of the resized tube-level mask prediction from the previous stage, following Cheng~\etal~\cite{cheng2021mask2former}. 
Instead of considering only thing masks as in previous works~\cite{VIS_TR,seqformer}, we jointly process both thing and stuff masks, shown in Equation~\eqref{equ:sp_attention}. 
$\rm{MLP}$ denotes linear layers to transform the object query, while $\mathbf{K}_{l}$ and $\mathbf{V}_{l}$ represent the spatial-temporal features transformed from $\mathbf{F}_{i}$, where $\mathbf{K}_{l}=\mathrm{Key}(\mathbf{F}_{i})$ and $\mathbf{V}_{l}=\mathrm{Value}(\mathbf{F}_{i})$, $\mathrm{Key}$ and $\mathrm{Value}$ are linear functions as in the common attention design. In the implementation, $F_{i}$ is sampled from the multi-scale feature output following Mask2Former, and we use the feature with the highest resolution for simple formulation purposes.

Within each tube, the query index is naturally the tracking ID for each object with no tracking association within the tube. 
This process is shown in the dash box area 
 of the Fig.~\ref{fig:method}.

\begin{figure}[t!]
	\centering
	\includegraphics[width=1.0\linewidth]{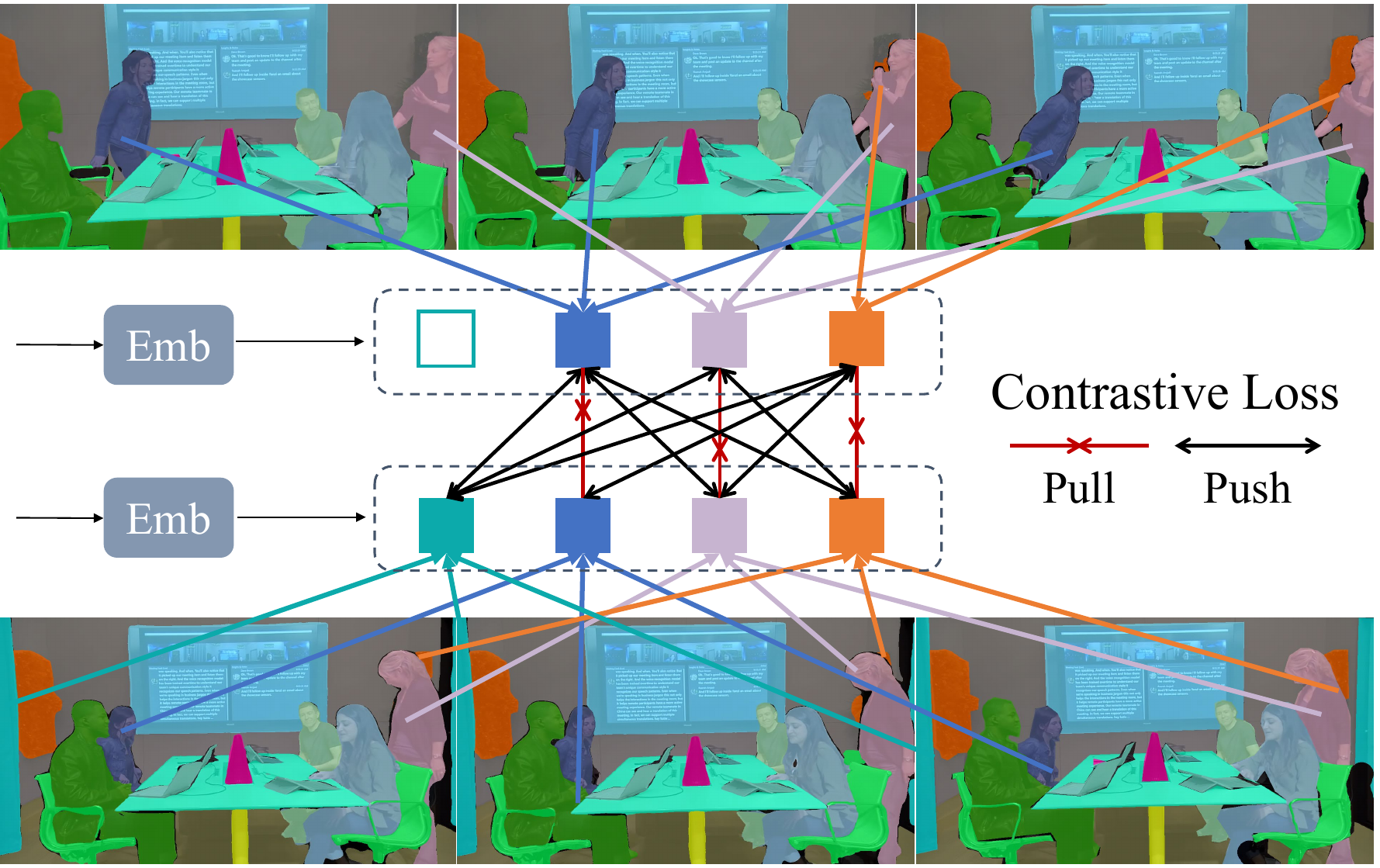}
	\caption{\small Illustration of the proposed cross-tube temporal contrastive learning. The input queries are first sent to the $\mathrm{Emb}$. Then cross-tube contrastive learning is performed. The same instance across different frames is indicated in the same color. We use spatial-temporal ground truth masks to perform positive/negative assignments to each query embedding.}
	\label{fig:method_cl}
\end{figure}

\noindent
\textbf{Cross-Tube Temporal Contrastive Learning.} Recent studies~\cite{li2022videoknet,IDOL,qdtrack} have demonstrated the effectiveness of contrastive learning in video segmentation. However, all of these studies perform learning on \textit{only two adjacent frames}, which is not suitable for tube-level matching. To capture a larger temporal context, we propose cross-tube temporal contrastive learning. 
While temporal contrastive learning is not new, our study is the first attempt to perform contrastive learning at the tube level.

As shown in Fig.~\ref{fig:method}, given a pair of subclips $v_{i}$ and $v_{j}$ as inputs, we first perform tube-level inference to obtain the global queries, $\mathbf{Q}_{i}$ and $\mathbf{Q}_{j}$, corresponding to two tubes. 
Note that both queries already encapsulate spatial-temporal information through Equation~\eqref{equ:sp_attention}. We randomly select two subclips from the neighborhood of all subclips, assuming that they contain corresponding objects (i.e., objects with the same IDs). We then add an extra lightweight embedding head $\mathrm{Emb}$ after each global query to learn the association embedding, which is implemented through several fully-connected layers. Following Li~\etal~\cite{li2022videoknet}, we use a mask-based assignment for contrastive learning.

Different from previous frame-wise methods, we propose a tube-wise label assignment strategy to form contrastive targets. 
Recall that our query embedding encodes information from more than one single frame, we use spatial-temporal masks (the same instance masks within the tube) for mask-based assignment, as shown in Fig.~\ref{fig:method_cl}. 
Specifically, we define a query embedding as positive to one object if its corresponding \textit{tube mask} has an IoU higher than $\alpha_1$, and negative if the IoU is lower than $\alpha_2$. We set $\alpha_1$ and $\alpha_2$ as 0.7 and 0.3, respectively. 
We use a sparse set of matched global queries for learning, where the query indices are assigned from ground truth tube masks.

%
We assume that there are $X$ matched queries from $\mathbf{Q}_{i}$ and $Y$ matched queries from $\mathbf{Q}_{j}$ as contrastive targets, where both $X$ and $Y$ are \textit{much fewer than} all queries $N$. 
The cross-tube temporal contrastive loss is written as:
\begin{align}
    \label{equ:qd_loss}
    L_\text{track} & = -\sum_{\textbf{y}^{+}}\text{log}
    \frac{\text{exp}(\textbf{x} \cdot \textbf{y}^{+})}
    {\text{exp}(\textbf{x} \cdot \textbf{y}^{+}) + \sum_{\textbf{y}^{-}}\text{exp}(\textbf{x} \cdot \textbf{y}^{-})},
\end{align}
where $\textbf{x}$, $\textbf{y}^{+}$, $\textbf{y}^{-}$ are query embeddings of tube pairs, their positive targets, and negative targets, which are sampled from $\mathbf{Q}_{i}$ and $\mathbf{Q}_{j}$, respectively, as illustrated in Fig.~\ref{fig:method}. 
The loss pulls positive embeddings close to each other and pushes the negative away, as shown in Fig.~\ref{fig:method_cl}. In addition, following previous work~\cite{qdtrack,li2022videoknet}, we also adopt L2 loss as an auxiliary loss to regularize the global query association process.

\begin{equation}
    \label{equ:qd_loss_aux}
    L_\text{track\_aux} = (\frac{\textbf{x} \cdot \textbf{y}}{||\textbf{x}|| \cdot ||\textbf{y}||} - b)^2,
\end{equation}
where $b$ is 1 if there is a match between the two samples, and 0 otherwise. Compared with previous works~\cite{li2022videoknet, IDOL} for contrastive learning, our loss considers additional temporal information, thus achieving a much better result. Experiments are reported in Sec.~\ref{sec:ablation}.

\noindent
\textbf{Cross-Tube Linking.} Apart from the improved supervision at the tube level, we take a further step to link tubes via their global queries, $\textbf{Q}_{i}$ and $\textbf{Q}_{j}$ for both training and inference. This encourages the interactions among tubes along the temporal dimension. 
We adopt a Multi-Head Self Attention ($\mathrm{MHSA}$) layer with a Feed Forward Network ($\mathrm{FFN}$)~\cite{vaswani2017attention} to learn the correspondence among each query to obtain the updated queries, allowing full correlation among queries. 
This process is shown as follows: 
\begin{equation}
    \mathbf{Q}_{j}^{f} = \mathrm{FFN}(\mathrm{MHSA}(\mathrm{Query}(\mathbf{Q}_{j}),\mathrm{Key}(\mathbf{Q}_{i}),\mathrm{Value}(\mathbf{Q}_{i})).
 \label{equ:selfattention}
\end{equation}
In this way, the information from the $i$-th tube is propagated to the $j$-th tube via the affinity matrix. 
The linked output $\mathbf{Q}_j^{f}$ is employed as the input to the embedding head $\mathrm{Emb}$, shown in the \textit{middle} of Fig.~\ref{fig:method} and the left of the Fig.~\ref{fig:method_cl}.

\begin{table}[!t]
	\centering
	\caption{\small System-level comparison between Tube-Link with related approaches~\cite{li2022videoknet,kim2022tubeformer,IDOL}.}
	\label{tab:comparison}
  \scalebox{0.62}{
    \begin{tabular}{ r c c c c}
    \toprule[0.15em]
     Property and Settings & Video K-Net~\cite{li2022videoknet} & TubeFormer~\cite{kim2022tubeformer} & IDOL~\cite{IDOL} & Our Tube-Link \\
    \toprule[0.15em]
    Online & \checkmark & \checkmark & \checkmark & \checkmark (n=1) \\
    Nearly Online &  & \checkmark &   &  \checkmark \\ 
    \hline
     VIS  & \checkmark & \checkmark & \checkmark & \checkmark \\
     VSS  & \checkmark &  \checkmark &  & \checkmark\\
     VPS  &  \checkmark & \checkmark &  & \checkmark \\
     \hline
     Mulitple Frames & \checkmark & &  \checkmark & \checkmark \\
     Frame Query Matching &  \checkmark & & \checkmark &\\
     Mask Mathicing  &  & \checkmark &  & \\
     Global Query Matching &  &  &  & \checkmark \\
    \bottomrule[0.2em]
    \end{tabular}
}
\end{table}

\noindent
\textbf{Relation with Previous Works.} 
We summarize the differences and properties of Tube-Link with previous methods in Table~\ref{tab:comparison}. 
Compared to Tubeformer~\cite{kim2022tubeformer}, Tube-Link explores cross-tube relationships from a global query matching perspective, while Tubeformer adopts simple mask matching. 
A detailed experimental comparison can be found in Sec.~\ref{sec:exp}. 
Compared to IDOL~\cite{IDOL} and Video K-Net~\cite{li2022videoknet}, our method explores multiple-frame information and global query matching, which is more robust for complex scenes and temporal consistency. 
If $n=1$, Tube-Link degenerates into the na\"{i}ve online method. 
It is noteworthy that since we process the video by sampling $n$ frames as input, we can obtain a much faster inference speed than online approaches via more efficient GPU memory usage and parallelization (see Sec.~\ref{sec:vis_analysis}).

\subsection{Training and Inference of Tube-Link}
\label{sec:train_and_inference}
\noindent
\textbf{Training and Loss Function.} 
In addition to the tracking loss in Equation~\eqref{equ:qd_loss}, we also use tube-wise segmentation loss. 
Specifically, we obtain the tube masks by stacking the mask of the same instance from different frames.
This establishes a one-to-one mapping between the predicted tube-level mask and the ground-truth tube-level mask based on the masked-based matching cost~\cite{detr,cheng2021mask2former}.
The tube-level masks are obtained from global queries $\mathbf{Q}$ like Mask2Former~\cite{cheng2021mask2former}. The final loss function is given as $L = \lambda_{cls}L_{t\_cls} + \lambda_{ce}L_{t\_ce} + \lambda_{dice}L_{t\_dice} + \lambda_{track}L_{track} + \lambda_{aux}L_{track\_aux}$. Here, $L_{t\_ce}$ is the Cross Entropy (CE) loss for tube-level classification, and $L_{t\_ce}$ and $L_{t\_dice}$ are tube-level mask Cross Entropy (CE) loss and Dice loss~\cite{dice_loss,wang2019solo} for segmentation, respectively. 
$L_{track}$ and $L_{track\_aux}$ are the tracking losses.  For VSS, we remove the $L_{track}$ term.

\noindent
\textbf{Simplicity.} We intend to keep a simple framework, and thus we \textbf{\textit{do not}} use any extra tricks or auxiliary losses employed in previous works, such as extra semantic segmentation loss~\cite{wang2020maxDeeplab}, tube-level mask copy and paste~\cite{kim2022tubeformer}, joint image dataset and video dataset pre-training~\cite{seqformer, IDOL}, etc.

\noindent
\textbf{Inference.} 
Taking VPS as an example, we use Tube-Link to generate tube-level panoptic segmentation masks from each input. We use the query embeddings from the learned embedding head $\mathrm{Emb}$ as association features and feed them as input to Quasi-Dense Tracker~\cite{qdtrack} in a near-online manner. Note that we only track the preserved instance masks from the panoptic segmentation maps, and the matching process is performed at \textit{tube-level} between two global queries. Unlike previous studies~\cite{deepsort, wangUnitrack}, we do not use extra motion cues to help track across subclips. Instead, we only use simple query feature matching across subclips, which saves time for track matching compared to online methods. An advantage of Tube-Link is its flexible inference via different subclip size settings $n$. Enlarging $n$ can improve both inference speed and performance on most datasets. For complex and high-resolution inputs, we decrease $n$ by only utilizing information within the temporal vicinity. We perform detailed ablations in Sec.~\ref{sec:ablation}. We will provide more details on inference for other tasks (VIS, VSS) in the appendix.

\section{Experiments}
\label{sec:exp}






\subsection{Experimental Settings}
\noindent
\textbf{Dataset.} We conduct experiments on five video datasets: VIPSeg~\cite{miao2022large}, VSPW~\cite{miao2021vspw}, KITTI-STEP~\cite{STEP}, and YouTube-VIS-19/21~\cite{vis_dataset}. We mainly conduct experiments on VIPSeg due to its scene diversity and long-length clips. The training, validation, and test sets of VIPSeg contain 2,806/343/387 videos with 66,767/8,255/9,728 frames, respectively. Although VSPW and VIPSeg share the same video clips, the training details are different since they are different tasks. Please refer to the \textit{supplementary material} for other datasets.

\noindent
\textbf{Evaluation Metrics.} For the VPS task, we adopt two metrics: $VPQ$~\cite{kim2020vps} and $STQ$~\cite{STEP}. The metric $STQ$ contains geometric mean of two items: Segmentation Quality ($SQ$) and Association Quality ($AQ$), where $ STQ = (SQ \times AQ)^{\frac{1}{2}}$. The former evaluates the pixel-level tracking, while the latter evaluates the pixel-level segmentation results in a video clip. For the VSS task, the Mean Intersection over Union (\textit{mIoU}) and mean Video Consistency ($mVC$)~\cite{miao2021vspw} are used for reference. For the VIS task, \textit{mAP} is adopted.

\noindent
\textbf{Implementation Details and Baselines.} We implement our models in PyTorch~\cite{pytorch_paper} with the MMDetection toolbox~\cite{chen2019mmdetection}. We use the distributed training framework with 16 V100 GPUs. Each mini-batch has one image per GPU. Following previous work, we use the image baseline pre-trained on COCO dataset~\cite{coco_dataset}. ResNet~\cite{resnet}, STDC~\cite{STDCNet}, and Swin Transformer~\cite{liu2021swin} are adopted as the backbone networks, which are pre-trained on ImageNet, and the remaining layers adopt the Xavier initialization~\cite{xavier_init}. 
For the detailed settings of other datasets, pretraining, and fine-tuning, please refer to the \textit{supplementary material}. To further verify the effectiveness of our approach, we build a stronger baseline by unifying Video K-Net with Mask2Former, where we replace the image encoder with Mask2Former. We term it Video K-Net+. We denote the extended Mask2Former-VIS for VPS as Mask2Former-VIS+.

\begin{table}[!t]
	\centering
	\caption{\small \textbf{Results on VIPSeg-VPS~\cite{miao2022large} validation dataset.} We report VPQ and STQ for reference. Following Miao~\etal~\cite{miao2022large}, we report VPQ scores at different window sizes (1, 2, 4, 6). We report the results obtained from either an efficient or a strong backbone for comparison.}
	\label{tab:vipseg_results}
  \scalebox{0.65}{
    \begin{tabular}{ r|c|cccccc}
    \toprule[0.15em]
     Method& backbone & $VPQ^{1}$ & $VPQ^{2}$ & $VPQ^{4}$ & $VPQ^{6}$ & VPQ & STQ \\
    \toprule[0.15em]
    VIP-DeepLab~\cite{ViPDeepLab} & ResNet50 & 18.4 & 16.9 & 14.8 & 13.7 & 16.0 & 22.0 \\
    VPSNet~\cite{kim2020vps} & ResNet50 & 19.9 & 18.1 & 15.8 & 14.5 & 17.0 & 20.8 \\
    SiamTrack~\cite{woo2021learning_associate_vps} & ResNet50 & 20.0 & 18.3 & 16.0 & 14.7 & 17.2 & 21.1 \\
    Clip-PanoFCN~\cite{miao2022large} & ResNet50 & 24.3 & 23.5 & 22.4 & 21.6 & 22.9 & 31.5 \\
    Video K-Net~\cite{li2022videoknet} & ResNet50 & 29.5 & 26.5 & 24.5 & 23.7 & 26.1 & 33.1 \\
    Video K-Net+~\cite{cheng2021mask2former,li2022videoknet} & ResNet50 & 32.1 & 30.5 & 28.5 & 26.7 & 29.1 & 36.6  \\
    Video K-Net~\cite{li2022videoknet} & Swin-base & 43.3 & 40.5 & 38.3 & 37.2 & 39.8 & 46.3 \\
    \hline
    Tube-Link & STDCv1 & 32.1 & 31.3 & 30.1 & 29.1 & 30.6 & 32.0 \\
    Tube-Link & STDCv2 & 33.2  & 31.8 & 30.6 & 29.6  &  31.4 & 32.8 \\
    \hline
    Tube-Link & ResNet50 & 41.2 & 39.5  & 38.0 & 37.0 &  39.2 & 39.5 \\
    Tube-Link & Swin-base & 54.5 & 51.4 & 48.6 & 47.1 & 50.4 & 49.4 \\
    \bottomrule[0.2em]
    \end{tabular}
}
\end{table}

\begin{table}[t]
  \centering
   \caption{\small \textbf{Results on the YouTube-VIS datasets.} We report the mAP metric. \textdagger~adopt COCO video pseudo labels. Axial means using the extra Axial Attention~\cite{axialDeeplab}. Our method does not apply these techniques for simplicity.}
  \label{tab:ytvis}
  \scalebox{0.68}{
  \begin{tabular}{l c | c  | c }
    \toprule[0.2em]
    Method & Backbone  & YTVIS-2019 & YTVIS-2021 \\
    \toprule[0.2em]
VISTR~\cite{VIS_TR} & ResNet50 & 36.2 & -  \\
TubeFormer~\cite{kim2022tubeformer} & ResNet50 + Aixal & 47.5  & 41.2  \\
IFC~\cite{hwang2021video} & ResNet50 & 42.8 & 36.6 \\
SeqFormer~\cite{seqformer} & ResNet50 & 47.4 & 40.5  \\
Mask2Former-VIS~\cite{cheng2021mask2former_vis}& ResNet50 & 46.4 & 40.6 \\
IDOL~\cite{IDOL} & ResNet50 & 46.4 & 43.9\\
IDOL~\cite{IDOL} \textdagger & ResNet50 & 49.5 & -\\
VITA~\cite{heo2022vita} \textdagger & ResNet50 & 49.8 & 45.7  \\
Min-VIS~\cite{huang2022minvis} &ResNet50& 47.4 & 44.2 \\
\hline
Tube-Link & ResNet50 & 52.8 & 47.9  \\
\hline
SeqFormer~\cite{seqformer} & Swin-large  & 59.3 & 51.8 \\
Mask2Former-VIS~\cite{cheng2021mask2former_vis} & Swin-large &  60.4 & 52.6 \\
IDOL~\cite{IDOL}  & Swin-large  & 61.5 & 56.1 \\ 
IDOL~\cite{IDOL}  & Swin-large \textdagger  & 64.3 & -\\
VITA~\cite{heo2022vita} \textdagger & Swin-large & 63.0 & 57.5 \\ 
Min-VIS~\cite{huang2022minvis} & Swin-large & 61.6 & 55.3 \\
\hline
Tube-Link & Swin-large  & 64.6 & 58.4  \\
    \bottomrule[0.2em]
  \end{tabular}
}
\end{table}

\begin{table}[t]
  \centering
    \caption{\small \textbf{Results on VSPW-VSS validation set}. $mVC_{c}$ means that a clip with $c$ frames is used.}
    \label{tab:vspw}
  \scalebox{0.68}{
  \begin{tabular}{l c c c c c }
    \toprule[0.2em]
    \textbf{VPSW} & Backbone & mIoU & $mVC_{8}$ &$mVC_{16}$  \\
    \toprule[0.2em]
    DeepLabv3+~\cite{deeplabv3plus} & ResNet101 & 35.7 & 83.5 & 78.4 \\
    TCB(PSPNet)~\cite{miao2021vspw,zhao2017pyramid} & ResNet101 & 37.5 & 86.9 & 82.1  \\
    Video K-Net (Deeplabv3+)~\cite{li2022videoknet,deeplabv3plus} & ResNet101  & 37.9 & 87.0 & 82.1 \\
    Video K-Net (PSPNet)~\cite{li2022videoknet,zhao2017pyramid} & ResNet101  & 38.0 & 87.2  & 82.3 \\
    MRCFA~\cite{sun2022mining} & MiT-B5 & 49.9 & 90.9  &  87.4  \\
    CFFM~\cite{sun2022vss} & MiT-B5 & 49.3 & 90.8 & 87.1 \\
    TubeFormer~\cite{kim2022tubeformer} & Axial-ResNet50x64  &  63.2 &  92.1 & 88.0 \\
    \hline
    Tube-Link & ResNet50 & 42.3 & 86.8 & 83.2 \\
    Tube-Link & Swin-large & 59.7 & 90.3 & 88.4 \\
    \bottomrule[0.2em]
  \end{tabular}
  }

\end{table}

\begin{table}[t]
  \centering
    \caption{\small \textbf{Results on VIP-Seg-VSS validation set}. $mVC_{c}$ means that a clip with $c$ frames is used.}
    \label{tab:vipseg_vss}
  \scalebox{0.68}{
  \begin{tabular}{l c c c c c }
    \toprule[0.2em]
    \textbf{VPSW} & Backbone & mIoU & $mVC_{8}$ &$mVC_{16}$  \\
    \toprule[0.2em]
    Video K-Net (Deeplabv3+)~\cite{li2022videoknet,deeplabv3plus} & ResNet101  & 38.3 & 88.0 & 83.1 \\
    Video K-Net (PSPNet)~\cite{li2022videoknet,zhao2017pyramid} & ResNet101  & 39.0 & 88.2  & 84.2 \\
    Mask2Former~\cite{cheng2021mask2former} &  ResNet50 & 38.4 & 87.5 & 82.5 \\
    Video K-Net+~\cite{cheng2021mask2former,li2022videoknet} &  Swin-base & 57.2 & 90.1 & 87.8  \\
    \hline
    Tube-Link & ResNet50 & 43.4 & 89.2 & 85.4 \\
    Tube-Link & Swin-base & 62.3 & 91.4 & 89.3 \\
    Tube-Link & Swin-large & 64.9 & 92.4 & 89.9 \\
    \bottomrule[0.2em]
  \end{tabular}
  }

\end{table}

\subsection{Benchmark Results}

\begin{table}[t]
  \centering
   \caption{\small \textbf{Results on the KITTI val set.} OF refers to an optical flow network~\cite{teed2020raft}.}
  \label{tab:kitti_step}
  \scalebox{0.68}{
  \begin{tabular}{l c c || c c c c }
    \toprule[0.2em]
    \textbf{KITTI-STEP} & Backbone & OF & STQ & AQ & SQ & VPQ \\
    \toprule[0.2em]
    P + Mask Propagation & ResNet50 & \checkmark & 0.67 & 0.63 & 0.71 & 0.44 \\
    Motion-Deeplab~\cite{STEP}& ResNet50 &  & 0.58 & 0.51 & 0.67 & 0.40  \\
    VPSNet~\cite{kim2020vps}& ResNet50  & \checkmark & 0.56 & 0.52 & 0.61 & {0.43}  \\
    TubeFormer-DeepLab~\cite{kim2022tubeformer} & ResNet-50 + Axial &  & 0.70 & 0.64 &  0.76 & 0.51 \\
    Video K-Net~\cite{li2022videoknet} & ResNet50 &  & 0.71 & 0.70  & 0.71  &  0.46 \\
    Video K-Net~\cite{li2022videoknet} & Swin-base &  & 0.73 & 0.72 & 0.73 & 0.53 \\
    \hline
    Tube-Link & ResNet50 &  & 0.68 & 0.67 & 0.69 & 0.51 \\
    Tube-Link & Swin-base &  & 0.72 & 0.69 & 0.74 & 0.56 \\
    \bottomrule[0.2em]
  \end{tabular}
  }
  \vspace{-4mm}
\end{table}

\begin{figure}[t]
  \centering
   \includegraphics[width=0.80\linewidth]{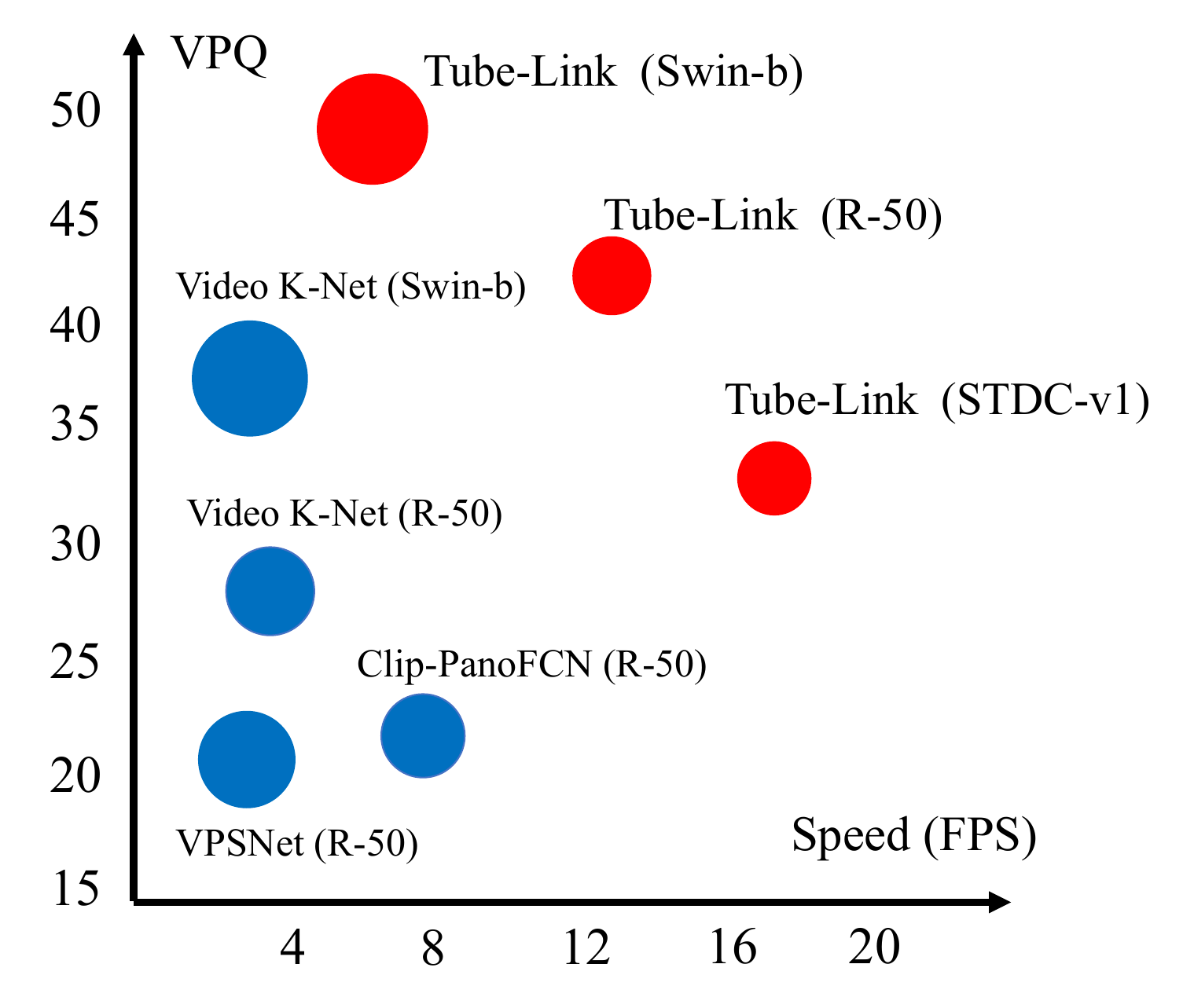}
   \caption{\small Tube-Link also achieves the best accuracy and speed trade-off on VIP-Seg dataset. FPS is measured on RTX GPU.}
   \label{fig:curve_trade_off_vipseg}
\end{figure}

\noindent
\textbf{[VPS] Results on VIPSeg.} 
We present the results of our Tube-Link method compared to previous works on the VIPSeg dataset in Tab.~\ref{tab:vipseg_results}. Our approach outperforms Video K-Net\cite{li2022videoknet} (under the same backbone) with 12\%-15\% VPQ and 7\%-10\% STQ improvements, respectively. Notably, our method with Swin-base~\cite{liu2021swin} backbone achieves new state-of-the-art results. 
We also evaluate our method using a lightweight backbone~\cite{STDCNet} for more efficient inference on video clips, and it achieves even better results than all previous methods with a larger ResNet50 backbone. 
These results demonstrate the effectiveness of our approach in exploiting temporal information.  Benefiting from the joint inference of subclips, our method achieves a much faster inference speed, as shown in Fig.~\ref{fig:curve_trade_off_vipseg}.

\begin{table*}[h!]
    \footnotesize
	\centering
	\caption{\small \textbf{Ablation studies and comparative analysis on VIPSeg validation set with the ResNet50 backbone.} 
	}
    \subfloat[Ablation Study on Each Component.]{
    \label{tab:ablation_a}
	    \begin{tabularx}{0.43\textwidth}{c c c c c} 
		        				\toprule[0.15em]
    	baseline  & TCL & CTL & $\mathrm{VPQ_{th}}$ & VPQ \\
        \toprule[0.15em]
            Mask2Former-VIS+ (F) & - & - & 29.4 & 32.4 \\
            \hline
            Mask2Former-VIS+ (T) & - & - & 31.0 & 34.5\\
             & \checkmark & - & 34.6  & 36.8  \\  
          \rowcolor{gray!15}  & \checkmark & \checkmark & 35.1 & 37.5 \\  
        \bottomrule[0.1em]
	    \end{tabularx}
    } \hfill
    \subfloat[Design Choices of TCL.]{
    \label{tab:ablation_b}
		\begin{tabularx}{0.28\textwidth}{c c c} 
			\toprule[0.15em]
			Method & VPQ & STQ \\
			\midrule[0.15em]
            Dense Query~\cite{qdtrack} & 30.2  & 30.1  \\
            Sparse Query~\cite{li2022videoknet} & 34.5  & 35.1 \\
            \rowcolor{gray!15} Global Query(Ours) &  37.5  & 36.5 \\
			\bottomrule[0.1em]
		\end{tabularx}
    } \hfill
    \subfloat[Association Target Assign.]{
    \label{tab:ablation_c}
		\begin{tabularx}{0.24\textwidth}{c c c} 
			\toprule[0.15em]
			Method & VPQ & STQ  \\
			\midrule[0.15em]
			All-Masks~\cite{qdtrack} & 30.1 & 29.2 \\
			GT-Mask~\cite{li2022videoknet} & 35.6 & 35.9 \\
			\rowcolor{gray!15} Tube-Mask & 37.5 & 36.5 \\
			\bottomrule[0.1em]
		\end{tabularx}
    } \hfill
    \vspace{2mm}
    \subfloat[Input Sub-clip Size with Tube Window Size of 2 as Input.]{
     \label{tab:ablation_d}
	    \begin{tabularx}{0.30\textwidth}{c c c c} 
		        				\toprule[0.15em]
    		 Clip Size & STQ & VPQ & $\mathrm{VPQ_{th}}$  \\
    		\toprule[0.15em]
    	    T=1 & 34.5 & 35.6 & 30.2 \\
    	    \rowcolor{gray!15} T=2 & 36.5 & 37.5 & 35.1 \\
    	    T=2(ovl) & 35.9 & 37.3 & 35.0 \\
    	    T=3 &  36.4 & 37.0 & 35.3 \\
        	\bottomrule[0.1em]
	    \end{tabularx}
    } \hfill
    \subfloat[Tube-Window for Inference with Input Sub-clip Size 2 for Training.]{
     \label{tab:ablation_e}
	    \begin{tabularx}{0.30\textwidth}{c  c c c} 
		        				\toprule[0.15em]
    		 Window Size & STQ & VPQ  & $\mathrm{VPQ_{th}}$ \\
    		\toprule[0.15em]
    	    W=2 &  36.5 & 37.5 & 35.1 \\
    	    W=4 &  39.2 & 39.0 & 38.2 \\
    	   \rowcolor{gray!15} W=6 &  39.5 & 39.2 & 38.9 \\
    	    W=8 &  38.3 & 38.5 & 37.3 \\
        	\bottomrule[0.1em]
	    \end{tabularx}
    } \hfill
    \subfloat[Tracking Choices with the Default Setting of Tab.(d). ]{
     \label{tab:ablation_f}
	    \begin{tabularx}{0.35\textwidth}{c c c c} 
		        				\toprule[0.15em]
    		 Settings  &  STQ & VPQ & $\mathrm{VPQ_{th}}$ \\
    		 \toprule[0.15em]
    		  Extra Tracker~\cite{wangUnitrack,deepsort}& 33.9 & 36.6 & 34.1 \\
    		  RoI Features~\cite{qdtrack} & 34.5 & 35.9 & 34.5 \\
    		  Query Embedding~\cite{li2022videoknet}  & 33.1  & 36.0  & 33.0 \\
    	     \rowcolor{gray!15} Our Tube embedding & 36.5 & 37.5 & 35.1\\
        	\bottomrule[0.1em]
	    \end{tabularx}
    } \hfill
\end{table*}

\noindent
\textbf{[VIS] Results on YouTube-VIS-2019/2021.} In Tab.~\ref{tab:ytvis}, we compare our method with state-of-the-art VIS methods on the YouTube-VIS 2019 and 2021 datasets. Our method achieves a 3.0\% and 2.2\% mAP gain over VITA~\cite{heo2022vita} when using the ResNet50 backbone. Furthermore, compared with the Mask2Former-VIS baseline~\cite{cheng2021mask2former_vis}, our method achieves 4-5\% mAP gains on the two datasets with different backbones. Our method also outperforms the previous near-online method TubeFormer~\cite{kim2022tubeformer} by 5-6\% in terms of mAP on the two VIS datasets.

\noindent
\textbf{[VSS] Results on VSPW and VIP-Seg.} We further conduct experiments on VSPW dataset~\cite{miao2021vspw} for VSS to demonstrate the generalization of Tube-Link. As shown in Tab.~\ref{tab:vspw}, our method achieves over 4\% mIoU improvement compared to the Mask2Former baseline. Under the same ResNet101 backbone, our method achieves the best results. Using the Swin base backbone, our method achieves about 3.7\% mIoU gains over Video K-Net+ with consistent improvements on $mVC$. Our method with a lightweight backbone achieves comparable results to DeepLabv3+ with ResNet101, but with about four times faster inference speed (shown in Fig.~\ref{fig:curve}). Without using any additional techniques, our method also outperforms recent methods specifically designed for VSS~\cite{sun2022vss,sun2022mining}. In Tab.~\ref{tab:vipseg_vss}, we also compare the video semantic segmentation methods in recent VIPSeg datasets with higher-resolution images. Compared with previous state-of-the-art methods, our approaches also achieve state-of-the-art results.


\noindent
\textbf{[VPS] Results on KITTI STEP.} 
We further validate our method on KITTI STEP~\cite{STEP} and report the results in Tab.~\ref{tab:kitti_step}. Our method achieves 0.51 VPQ with the ResNet50 backbone, setting a new state-of-the-art result \textit{without} using temporal attention or optical flow warping. When using a strong Swin-base~\cite{liu2021swin} backbone, our method still achieves better results than Video K-Net~\cite{li2022videoknet} by 3\% VPQ and comparable results on STQ. It is worth noting that one can further improve the performance of Tube-Link by employing a better tracker design.

\subsection{Ablation Study and Visual Analysis}
\label{sec:ablation}


\noindent
\textbf{Improvements over Strong VPS Baseline.} 
In Tab.~\ref{tab:ablation_a}, we demonstrate the effectiveness of each component proposed in Sec.~\ref{sec:tb_framework}. 
The first row shows the results of the frame matching baseline. After adopting the tube matching, we obtain a gain of 1.6\% $\mathrm{VPQ_{th}}$ and 2.1\% on VPQ, even without any specific tracking design, which results in the same observation as shown in Tab.~\ref{tab:toy_exp}. Thus, we use Mask2Former-VIS+ (T, T=2) as our baseline by default, which achieves a strong starting point of 34.5 VPQ. $\mathrm{VPQ_{th}}$ refers to the VPQ for the thing class. This result shows the effectiveness of the na\"{i}ve framework. The addition of TCL further boosts performance, with a gain of 3.5\% on $\mathrm{VPQ_{th}}$ and 1.7\% on VPQ. Furthermore, adding CTL, which makes the association more consistent, improves $\mathrm{VPQ_{th}}$ by 1.5\%.

\noindent
\textbf{Ablation on Temporal Contrastive Loss.} We also compare our TCL design with previous works that use dense queries~\cite{qdtrack} or sparse queries~\cite{li2022videoknet} for matching. Both settings use only one frame, while our subclip size is two. As shown in Tab.~\ref{tab:ablation_b}, our method achieves the best results since tube matching encodes more temporal information. In particular, we observe 3.0\% VPQ improvements compared to the strong Video K-Net baseline.

\begin{figure}[t!]
	\centering
	\includegraphics[width=1.0\linewidth]{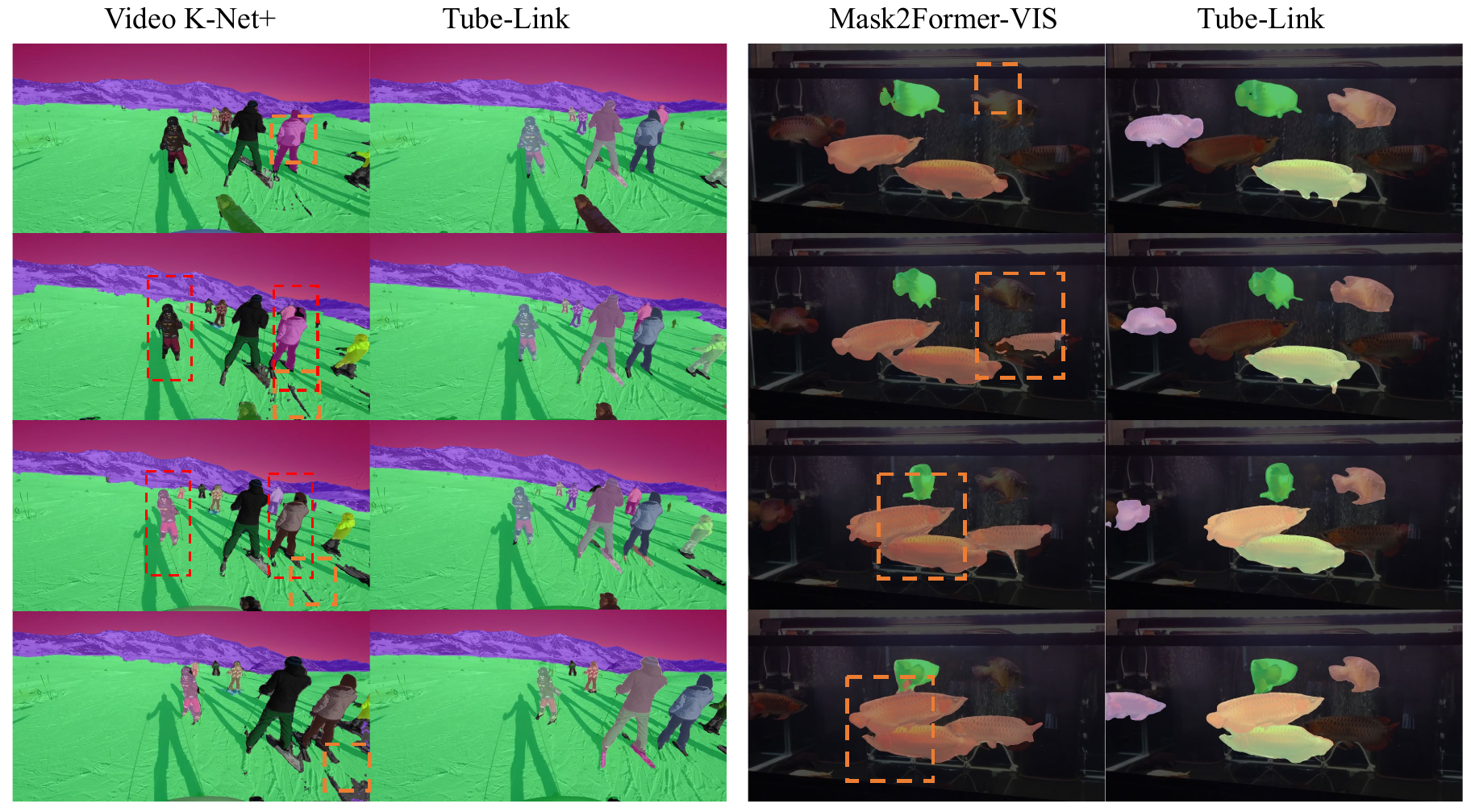}
	\caption{\small Comparison results on VIP-Seg and YuoTube-VIS. Our method achieves consistent segmentation (shown in orange boxes) and better tracking results (shown in red boxes).}
	\label{fig:visulize}
\end{figure}

\begin{figure}[t]
  \centering
   \includegraphics[width=1.\linewidth]{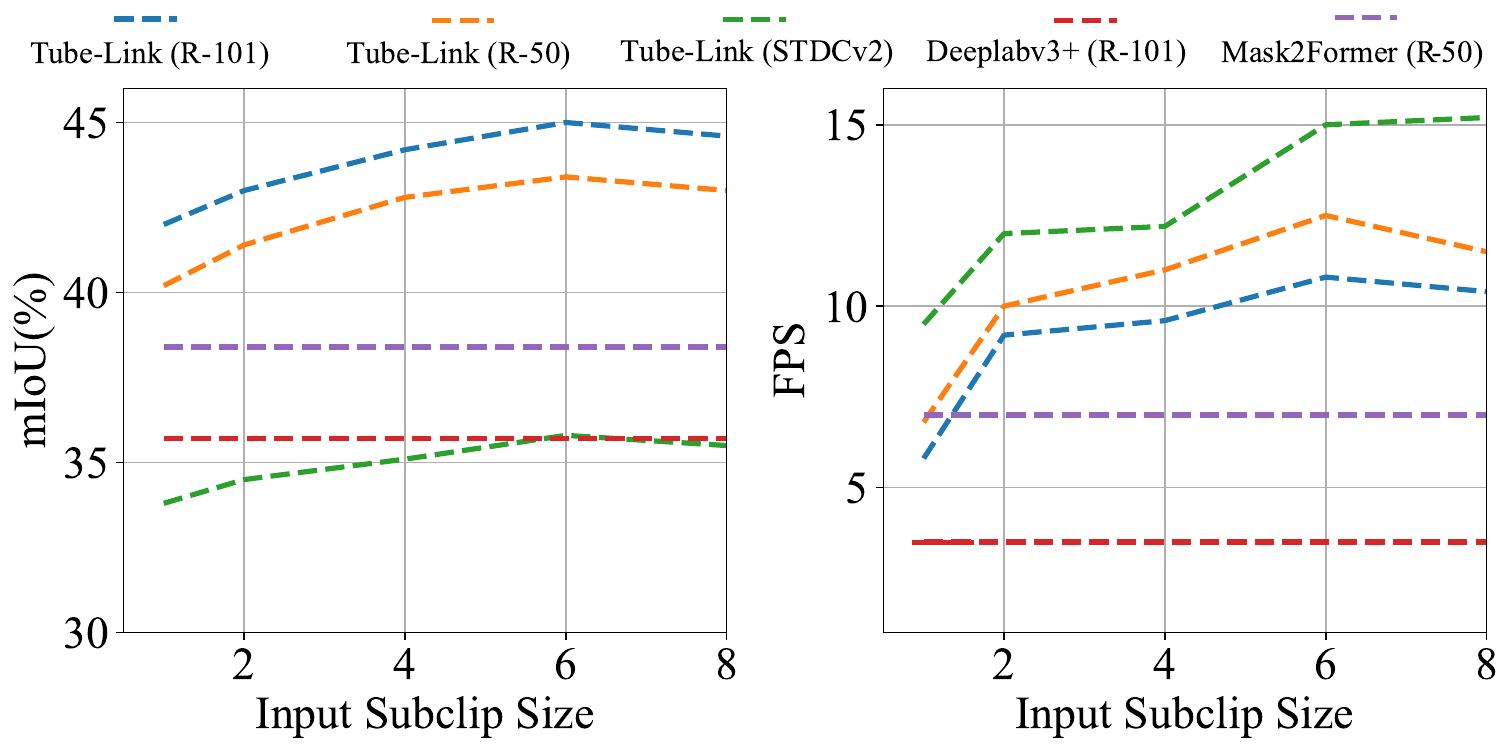}
   \caption{\small Efficiency Analysis of Tube-Link. Left: Segmentation results (mIoU) of VSPW with different subclip sizes. Right: Inference speed (FPS) with different subclip sizes.}
   \label{fig:curve}
\end{figure}

\noindent
\textbf{Ablation on Association Target Assignment.} 
In Table \ref{tab:ablation_c}, we show the results of the ablation study on building association targets. We find that using a tube-level mask achieves the best results. Using the mask from one of the input subclips leads to inferior results. This is because the ground truth masks of a single frame are not aligned with the input global queries, where the global queries are learned from multiple frames using Equation \eqref{equ:sp_attention}.

\noindent
\textbf{Effect of Sub-clip Size for Training.} 
In Tab.~\ref{tab:ablation_d}, we investigate the impact of subclip size on training. Tube-Link becomes an online method when the subclip size is 1. As shown in the table, enlarging the subclip size improves the performance. We also examine overlapping during sampling, denoted as ovl, where two input subclips overlap at one frame. As shown in Tab.~\ref{tab:ablation_d}, enlarging the subclip size to 2 achieves significant improvement. However, we find that either frame overlapping or using a larger subclip size ($T=3$) does not bring extra gains. Adding more frames does not benefit temporal association learning, since most instances are similar within a subclip. Moreover, using more frames is not memory-friendly during training. Thus, the subclip size is set to 2. We can enlarge the size for inference, as shown in Tab.~\ref{tab:ablation_e}.
  

\noindent
\textbf{Effect of Sub-clip Size for Inference.} 
%
During training, the subclip size is limited due to memory constraints, but we can expand it during inference to improve the performance. For instance, we use a subclip size of 2 during training and increase it to 6 during inference. Tab.~\ref{tab:ablation_e} shows that enlarging the subclip size for inference improves the performance considerably for all three metrics: STQ, VPQ, and $\mathrm{VPQ_{th}}$. However, when the subclip size is further increased to 8, the performance drops because the global queries are not designed to handle larger subclips. Increasing the subclip size can also speed up the inference process by utilizing the full GPU memory, as demonstrated in Fig.~\ref{fig:curve}.


\noindent
\textbf{Different Tracking Choices.} 
In Tab.~\ref{tab:ablation_f}, we compare different tracking approaches used in previous studies~\cite{qdtrack,li2022videoknet,deepsort} with our Tube-Link. Our Tube-Link only uses the learned tube-level embedding for the association. We find that the default tube embedding works best in our framework, without requiring any association embedding head or RoI crop operation on the VIPSeg dataset.  

\subsection{Visualization and More Analysis}
\label{sec:vis_analysis}

\noindent
\textbf{GFLops and Parameter Analysis.} Compared with Mask2Former baseline, we only add one $\mathrm{Emb}$ head and one self-attention layer, introducing only 2.2\% GFLops and 1.4\% extra parameters with $720 \times 1280$ input. 



\noindent
\textbf{Speed and Accuracy with Different Input Subclip Size.} 
As shown in Table \ref{tab:ablation_e}, adding more frames improves the VPS results. To further analyze the speed-accuracy trade-off, we present a detailed comparison of different methods on the VSPW dataset in Fig.~\ref{fig:curve}. The left plot shows that enlarging the subclip size also improves the VSS results. The right plot illustrates that increasing the subclip size improves the single-frame baseline by 1.25-1.5\% for various backbones. Both performance and speed reach a plateau when the size increases to 6. The experiment justifies our choice of using an input subclip size of 6 for inference.

\noindent
\textbf{Visual Improvements on Baselines.} In Fig.~\ref{fig:visulize}, we present the visual comparison with several strong baselines (Video K-Net+ and Mask2Fomer-VIS) in VPS and VIS settings. The results are randomly sampled from a long clip. We achieve better results on both segmentation and tracking. More visual examples can be found in the supplementary material. 

\section{Conclusion}
\label{sec:conclusion}

We present Tube-Link, a simple yet flexible universal video segmentation framework. Our key insight is to perform cross-tube matching rather than cross-frame matching. By equipping the Mask2Former architecture with the proposed cross-tube learning, Tube-Link achieves new state-of-the-art results in all three major video segmentation tasks (VSS, VIS, VPS) using one unified architecture.

\noindent
\textbf{Limitation and Future Work.}
Tube-Link is pre-trained on image datasets and requires re-training for each new video dataset. In the future, we hope to develop a model that can be trained only once to unify universal image and video segmentation tasks. One potential solution can be training our Tube-Link with the merged image and video datasets using CLIP~\cite{radford2021learning,wu2023open} text embedding to unify labels. Then, we can build a universal image/video model for various scenes.

\noindent
\textbf{Broader Impact.} Our work pushes the boundary of video panoptic segmentation algorithms in a simple, flexible, and efficient way. The proposed framework provides a unified and general solution for dense video segmentation, which has the potential to greatly simplify and expedite model development in various real-world applications that heavily rely on video input, including autonomous driving and robot navigation.

\noindent
\textbf{Acknowledgement.} This study is supported under the RIE2020 Industry Alignment Fund Industry Collaboration Projects (IAF-ICP) Funding Initiative, as well as cash and in-kind contribution from the industry partner(s). It is also supported by Singapore MOE AcRF Tier 1 (RG16/21).

\appendix
\section{Appendix}

\noindent
\textbf{Overview.} In addition to the main paper, we further list the following details and more experiment results as supplementary to our work.

\begin{enumerate}
\setlength{\leftmargin}{-1em}
\setlength{\parsep}{0ex} 
\setlength{\topsep}{0ex}
\setlength{\itemsep}{0.5ex}  
\setlength{\labelsep}{0.5em} 
\setlength{\itemindent}{-0.5em} 
\setlength{\listparindent}{0em} 
\item More detailed description and comparison of Tube-Link. (Sec.~\ref{sec:more_decription_tube_link})
\item Detailed experiment settings and implementation details for each dataset. (Sec.~\ref{sec:implemntation_details})
\item More ablation studies and experiment results.(Sec.~\ref{sec:more_ablation_and_exp_results}) 
\item More visual results. (Sec.~\ref{sec:vis_results})
\end{enumerate}

\begin{table*}[!t]
   \centering
    \caption{\small Different Setting Comparison with previous VIS and VPS methods.}
   \scalebox{0.65}{
   \setlength{\tabcolsep}{2.5mm}{\begin{tabular}{c | c c   c | c  c | c | c c c c}
      \toprule[0.15em]
        Method &  VSS & VIS & VPS & Online & Nearly Online & Joint Mulitple Frames  & Frame Matching & Tube Matching & Mask Matching & No Association (use Query Index) \\
        \hline 
        \rowcolor{gray!15} CFFM~\cite{sun2022vss} & \checkmark &  &  &  & \checkmark  & \checkmark &  &  &  & \checkmark \\
         \rowcolor{gray!15} MRCFA~\cite{sun2022mining} & \checkmark &  &  &  & \checkmark  & \checkmark &  & & & \checkmark  \\
         \hline
        \rowcolor{orange!15} Cross-VIS~\cite{CrossVIS} & &  \checkmark &  & \checkmark & & & \checkmark & & &\\
        \rowcolor{orange!15} IDOL~\cite{IDOL} & &  \checkmark &  & \checkmark & & & \checkmark & & &  \\
        \rowcolor{orange!15} SeqFormer~\cite{seqformer} & & \checkmark & &  & \checkmark & \checkmark & & & & \checkmark \\
         \rowcolor{orange!15}EfficientVIS~\cite{EfficientVIS}& & \checkmark & & & \checkmark & \checkmark & & & & \checkmark\\   
         \rowcolor{orange!15}VITA~\cite{heo2022vita} & & \checkmark & & & \checkmark & \checkmark & & & & \checkmark \\
        \rowcolor{orange!15} Min-VIS~\cite{huang2022minvis} & & \checkmark & & \checkmark & & & \checkmark & & & \\
         \rowcolor{orange!15} IFC~\cite{hwang2021video} & & \checkmark &   & & \checkmark & \checkmark &  &  \checkmark & & \\
         \rowcolor{orange!15} Gen-VIS~\cite{heo2022generalized} & & \checkmark & & \checkmark & \checkmark & \checkmark & & \checkmark & &  \\
        \hline
        \rowcolor{blue!15} SLOT-VPS~\cite{slot_vps} &  & & \checkmark &  & \checkmark & \checkmark & & & & \checkmark \\
        \rowcolor{blue!15} TubeFormer~\cite{kim2022tubeformer} &  \checkmark & \checkmark & \checkmark &  & \checkmark & \checkmark & & & \checkmark &   \\
        \rowcolor{blue!15} Video K-Net~\cite{li2022videoknet} & \checkmark & \checkmark & \checkmark & \checkmark &  &  & \checkmark & & & \\
        \rowcolor{red!15} Our Tube-Link & \checkmark & \checkmark & \checkmark & \checkmark & \checkmark & \checkmark &  & \checkmark &  & \\
      \bottomrule[0.10em]
   \end{tabular}}}
   \label{tab:more_detailed_comparison}
\end{table*}

\subsection{More Detailed Description of Tube-Link}
\label{sec:more_decription_tube_link}

This section presents the method details, including several baselines and Tube-Link inference for different datasets. 
Then, due to the limited pages in the main paper, we compare several closely related works in VIS and VPS in detail.

\vspace{2mm}
\noindent
\textbf{Video K-Net+ Baseline.} This baseline is based on two previous state-of-the-art methods, including Video K-Net~\cite{li2022videoknet} and Mask2Former~\cite{cheng2021mask2former}. 
In particular, Video K-Net is based on K-Net~\cite{zhang2021knet}, an image panoptic segmentation model. We replace K-Net with Mask2Former~\cite{cheng2021mask2former}, and the remaining parts are the same as the Video K-Net. 
Since the performance of Mask2Former is better than K-Net on image segmentation datasets, Video K-Net+ serves as the strong online baseline for both VPS and VSS tasks.

\vspace{2mm}
\noindent
\textbf{Detailed Inference Procedure of Panoptic Matching.} During the inference, we perform tube-level panoptic matching according to learned association embeddings only on final panoptic tube masks.
In particular, we save the index of each global query from the final panoptic tube results, and then we use these indexed queries via the embedding head $Emb$. 

\vspace{2mm}
\noindent
\textbf{Detailed Inference Procedure of VSS and VIS.} Since VSS does not need tracking, we do not apply the extra tracking embedding during the inference. 
Instead, the tube mask logits are obtained directly from the dot product between global queries and spatial-temporal decoder features. 
The final segmentation labels are directly obtained via argmax on predicted logits. 
For VIS, we follow nearly the same procedure as VPS, except no stuff queries are involved.

\vspace{2mm}
\noindent
\textbf{More Detailed Comparison with Previous Nearly Online Approaches in VIS and VPS.} In addition to the main paper, in Tab.~\ref{tab:more_detailed_comparison}, we present a more detailed comparison with previous works on VIS and VPS. From the table, our method uses tube-wised matching and supports all three video segmentation tasks in one architecture. 

In particular, both SLOT-VPS~\cite{slot_vps} and SeqFormer~\cite{seqformer} also adopt multiple frames design. However, there are no data association processes involved. Moreover, they are designed for VIS and VPS, individually, and our method outperforms the SeqFormer on two VIS datasets, as shown in Tab.~\ref{tab:ytvis_supp}. Furthermore, unlike SLOT-VPS and SeqFormer, our method can handle long video inputs.

Gen-VIS~\cite{heo2022generalized} also adopts the tube-wised design, which combines the nearly online method and online method in one framework. However, it can not support other video segmentation tasks, including VSS and VPS. Moreover, it is  
not verified in more complex scenes, including the driving dataset KITTI-STEP~\cite{STEP} and the recent more challenging dataset VIP-Seg~\cite{miao2022large}. In contrast, our Tube-Link is fully verified by three different video segmentation tasks and five different datasets. In particular, using the same ResNet50 backbone and detector~\cite{cheng2021mask2former}, even without COCO video joint training, our method works better than Gen-VIS~\cite{heo2022generalized}, as shown in Tab.~\ref{tab:ytvis_supp}.


\subsection{Implementation Details}
\label{sec:implemntation_details}

\noindent
\textbf{Detailed Training and Inference on VIP-Seg.} We use the COCO-pretrained model following~\cite{miao2022large}. The entire training process takes eight epochs. We adopt multiscale training where the scale ranges from 1.0 to 2.0 of the original image size, and then we apply a random crop of 720 $\times$ 720 patches. In particular, we perform the augmentation for each frame in the sampled subclips. For the inference, the subclip window size is set to six by default. We pad the remaining frames in the last subclip by repeating the last frame. We drop the padded results for evaluation.

\vspace{2mm}
\noindent
\textbf{Detailed COCO pretraining setting.} 
For COCO~\cite{coco_dataset} panoptic segmentation dataset pretraining, all the models are trained following original Mask2Former settings~\cite{zhang2021knet}. We adopt the multiscale training setting as previous work~\cite{detr} by resizing the input images such that the shortest side is at least 480 and 800 pixels, while the longest size is at most 1333. For data augmentation, we use the default large-scale jittering (LSJ) augmentation with a random scale sampled from the range 0.1 to 2.0 with the crop size of 1024 $\times$ 1024. For ResNet50~\cite{resnet} and Swin-base~\cite{liu2021swin} model, we train the model with 50 epochs following the original settings. For STDC model~\cite{STDCNet}, we train the model for 36 epochs.

\vspace{2mm}
\noindent
\textbf{Detailed Training and Inference on KITTI-STEP dataset.} For KITTI-STEP training, we follow previous Motion-Deeplab~\cite{STEP} and Video K-Net~\cite{li2022videoknet}, we adopt multiscale training where the scale ranges from 1.0 to 2.0 of origin images size. We then apply a random crop of 384 $\times$ 1248 patches. The total training epoch is set to 12. The inference procedure is the same as Cityscapes-VPS dataset. Following the previous works~\cite{STEP,li2022videoknet}, we also use Cityscapes dataset~\cite{cordts2016cityscapes} pretraining before training on STEP. Pretraining on Cityscapes STEP datasets further leads to 3\% VPQ and 2\% STQ improvements. We adopt the same inference pipeline as VIP-Seg, where we set the subclip window size to 2. We \textbf{do not} pre-train our model on the COCO dataset for a fair comparison.

\vspace{2mm}
\noindent
\textbf{Detailed Training and Inference on VSPW dataset.} We adopt nearly the same training pipeline for VSPW as VIP-Seg. The main difference is that we adopt longer training epochs, where we set the training epochs to 12, where we find about 1\% mIoU gain over different baselines. Moreover, we remove the tracking loss since we only focus on segmentation quality.

\vspace{2mm}
\noindent
\textbf{Detailed Training and Inference on Youtube-VIS-2019/2021 datasets.} We follow the same setting as Mask2Former-VIS~\cite{cheng2021mask2former_vis}. We train our models for 6k iterations, with a batch size of 16 for YouTubeVIS-2019 and 8k iterations for YouTubeVIS-2021.
All models are initialized with COCO instance segmentation models of Mask2Former. Different from previous SOTA VIS models~\cite{heo2022vita,seqformer,IDOL}, we only use YouTubeVIS training data, and \textit{do not} use COCO video images for data augmentation. Moreover, we also do not apply clip-wised copy-paste that is used in TubeFormer~\cite{kim2022tubeformer}. The same training procedure is adopted for the OVIS dataset as well.

\begin{table}[t]
  \centering
   \caption{\small \textbf{More Ablation on Tube-Wised Matching in Youtube-VIS dataset.}}
  \label{tab:tube_wisedmethod}
  \scalebox{0.90}{
  \begin{tabular}{l | c c  }
    \toprule[0.2em]
    \textbf{Settings} & Youtube-VIS-2019 & Youtube-VIS-2021 \\
    \toprule[0.2em]
     tube size=1 & 47.8 & 44.2 \\
     tube size=2 & 49.8 &  45.9  \\
     tube size=3 & 51.3 &  46.2  \\
     \rowcolor{gray!15} tube size=4 & 52.8  &  47.9  \\
     tube size=6 &  51.2 & 46.8  \\
    \bottomrule[0.1em]
  \end{tabular}
  }
\end{table}

\begin{table}[t]
  \centering
   \caption{\small \textbf{Ablation on Inference with Overlapped Frames.} We use the STDC-v1 backbone. The subclip window size is 6.}
  \label{tab:over_lap_window_size}
  \scalebox{0.90}{
  \begin{tabular}{l | c c c c }
    \toprule[0.2em]
    \textbf{Settings} & STQ & VPQ & SQ & FPS \\
    \toprule[0.2em]
     \rowcolor{gray!15} No Overlapping &  32.0 & 30.6 & 28.4 &  16.2 \\
     Overlapping=1 & 31.0  &  30.5 & 28.5 & 14.6 \\
     Overlapping=2 & 32.3  &  31.2 & 29.1 & 10.2 \\
     Overlapping=4 &  33.1  & 31.6  & 28.6 & 8.4 \\
    \bottomrule[0.1em]
  \end{tabular}
  }
\end{table}

\begin{table}[t]
  \centering
   \caption{\small \textbf{Ablation on Effect of COCO Pretraining.} We use the STDC-v1 backbone.}
  \label{tab:abl_coco_pretrain}
  \scalebox{0.90}{
  \begin{tabular}{l c | c c c }
    \toprule[0.2em]
    \textbf{Settings} & Method & STQ & VPQ & SQ \\
    \toprule[0.2em]
    w COCO pretrained & Video K-Net+ & 26.1 & 25.8  & 25.2 \\
   \rowcolor{gray!15} w/o COCO pretrained & Video K-Net+ & 12.4 & 12.4  & 18.3 \\
    \hline
    w COCO pretrained & Tube-Link & 32.0 & 30.6 & 28.4 \\
   \rowcolor{gray!15} w/o COCO pretrained & Tube-Link & 21.8 & 16.8 & 20.3 \\

    \bottomrule[0.1em]
  \end{tabular}
  }
\end{table}

\begin{table}[t]
  \centering
   \caption{\small \textbf{Ablation on Training Epochs.} We use the STDC-v1 backbone.}
  \label{tab:ablation_train_epoch}
  \scalebox{0.95}{
  \begin{tabular}{l | c c c }
    \toprule[0.2em]
    \textbf{Settings} & STQ & VPQ & SQ \\
    \toprule[0.2em]
    Epoch=4 & 29.2  & 28.1 & 26.5  \\
   \rowcolor{gray!15} Epoch=8 &  32.0 & 30.6 & 28.4 \\
    Epoch=12 & 31.6 & 30.8  & 29.1 \\
    \bottomrule[0.1em]
  \end{tabular}
  }
\end{table}

\begin{table*}[t]
  \centering
   \caption{\small \textbf{Detailed Results on the Youtube-VIS datasets (2019/2021).} We report the mAP metric. \textdagger~adopts COCO video pseudo labels~\cite{heo2022vita,heo2022generalized,heo2022vita}. Axial means using the extra Axial Attention~\cite{axialDeeplab}. Our method does not apply these techniques for simplicity.}
  \label{tab:ytvis_supp}
  \scalebox{0.95}{
  \begin{tabular}{l c | ccccc | cccccc }
    \toprule[0.2em]
    Method & Backbone  & \multicolumn{5}{c}{YTVIS-2019} & \multicolumn{5}{c}{YTVIS-2021} \\
    - & - &  AP        & AP$_{50}$ & AP$_{75}$ & AR$_1$    & AR$_{10}$ & AP        & AP$_{50}$ & AP$_{75}$ & AR$_1$    & AR$_{10}$ \\
    \toprule[0.2em]
VISTR~\cite{VIS_TR} & ResNet50 & 36.2 & 59.8 & 36.9 & 37.2 & 42.4 & - & - & - & - & - \\
EfficientVIS~\cite{EfficientVIS} & ResNet-50     & 37.9      & 59.7      & 43.0      & 40.3      & 46.6  & 34.0     & 57.5    & 37.3      & 33.8      & 42.5  \\
TubeFormer~\cite{kim2022tubeformer} & ResNet50 + Aixal & 47.5 & 68.7 & 52.1 & 50.2 & 59.0 & 41.2 & 60.4 & 44.7 & 40.4 & 54.0  \\
IFC~\cite{hwang2021video} & ResNet50 & 41.2      & 65.1      & 44.6      & 42.3      & 49.6                                                                   & 35.2      & 55.9      & 37.7      & 32.6      & 42.9 \\
Seqformer~\cite{seqformer} & ResNet50 & 47.4      & 69.8      & 51.8      & 45.5      & 54.8                                                                   & 40.5      & 62.4      & 43.7      & 36.1      & 48.1  \\
Mask2Former-VIS~\cite{cheng2021mask2former_vis}& ResNet50 & 46.4      & 68.0      & 50.0      & -         & -  & 40.6      & 60.9      & 41.8      & -         & -   \\
IDOL~\cite{IDOL} & ResNet50 & 46.4  & - & -   & -         & - & 43.9  & - & -   & -         & - \\
IDOL~\cite{IDOL} \textdagger & ResNet50 & 49.5 & - & -   & -         & -   & - & -   & -         & -   & -  \\
VITA~\cite{heo2022vita} \textdagger & ResNet50 & 49.8      & 72.6     & 54.5      & 49.4      & 61.0  & 45.7      & 67.4      & 49.5      & 40.9     & 53.6  \\
Min-VIS~\cite{huang2022minvis} &ResNet50& 47.4      & 69.0      & 52.1      & 45.7      & 55.7     & 44.2      & 66.0      & 48.1      & 39.2      & 51.7 \\
Cross-VIS~\cite{CrossVIS} & ResNet50 & 36.3      & 56.8      & 38.9      & 35.6      & 40.7   & 34.2      & 54.4      & 37.9      & 30.4      & 38.2 \\
VISOLO~\cite{VISOLO} & ResNet50 &  38.6      & 56.3      & 43.7      & 35.7      & 42.5    & 36.9      & 54.7      & 40.2      & 30.6      & 40.9  \\
GenVIS~\cite{heo2022generalized} & ResNet50 & {51.3}  & 72.0    & {57.8}  & {49.5}  & 60.0 & {46.3}  & 67.0 & {50.2} & 40.6 & 53.2 \\
\hline
Tube-Link & ResNet50 & 52.8 & 75.4 & 56.5 & 49.3 &59.9 & 47.9  & 70.0 & 50.2 & 42.3 & 55.2 \\ 
\hline
SeqFormer~\cite{seqformer} & Swin-large  & 59.3      & 82.1      & 66.4      & 51.7      & 64.4  & 51.8      & 74.6      & 58.2      & 42.8      & 58.1  \\
Mask2Former-VIS~\cite{cheng2021mask2former_vis} & Swin-large &  60.4      & 84.4      & 67.0      & -    & -   & 52.6      & 76.4      & 57.2      & -         & -     \\
IDOL~\cite{IDOL}  & Swin-large  & 61.5 & -   & -         & -   & -  & 56.1 & -   & -   & -   & -  \\ 
IDOL~\cite{IDOL}  & Swin-large \textdagger  & {64.3}      & {87.5}      & {71.0}      & 55.6      & 69.1 & 56.1      & 80.8      & 63.5      &  45.0      & 60.1 \\
VITA~\cite{heo2022vita} \textdagger & Swin-large & 63.0    & {86.9}      & 67.9      & {56.3}    & 68.1    & 57.5      & 80.6   & 61.0      & 47.7   & 62.6 \\ 
Min-VIS~\cite{huang2022minvis} & Swin-large & 61.6      & 83.3      & 68.6      & 54.8      & 66.6    & 55.3      & 76.6      & 62.0      & 45.9      & 60.8\\
\hline
Tube-Link & Swin-large  & 64.6 & 86.6 & 71.3 &  55.9 & 69.1 & 58.4 & 79.4  & 64.3 & 47.5 & 63.6 \\
    \bottomrule[0.2em]
  \end{tabular}
}
\end{table*}

\begin{table}[t]
  \centering
   \caption{\small \textbf{Results on the OVIS datasets.} We report the mAP metric. \textdagger~adopts COCO video pseudo labels. Axial means using the extra Axial Attention~\cite{axialDeeplab}. Our method does not apply these techniques for simplicity.}
  \label{tab:ovis}
  \scalebox{0.95}{
  \begin{tabular}{l c c c c  c}
    \toprule[0.2em]
    Method & AP   & AP$_{50}$ & AP$_{75}$ & AR$_1$   & AR$_{10}$ \\
    \toprule[0.2em]
    CrossVIS~\cite{CrossVIS} & 14.9      & 32.7          & 12.1          & 10.3          & 19.8 \\
    VISOLO~\cite{VISOLO} & 15.3      & 31.0          & 13.8          & 11.1          & 21.7  \\
    TeViT~\cite{TeViT}  & 17.4      & 34.9          & 15.0          & 11.2          & 21.8  \\
    VITA~\cite{heo2022vita} & 19.6      & 41.2          & 17.4          & 11.7          & 26.0 \\
    DeVIS~\cite{DeVIS} & 23.8      & 48.0          & 20.8          & -             & -     \\
    Min-VIS~\cite{huang2022minvis} & 25.0      & 45.5          & 24.0          & 13.9          & 29.7 \\
    IDOL~\cite{IDOL}  & 30.2      & 51.3          & 30.0          & 15.0          & 37.5 \\
    VITA~\cite{heo2022vita} \textdagger&  19.6 & 41.2  & 17.4 & 11.7 & 26.0 \\
\hline
Tube-Link & 29.5 & 51.5 &  30.2 & 15.5 & 34.5 \\
    \bottomrule[0.2em]
  \end{tabular}
}
\end{table}

\begin{table}[!t]
    \caption{More Experiment Results.}
    \begin{subtable}{.50\linewidth}
        \centering
        \footnotesize
        \caption{More results on ViP-Seg dataset.}
        \label{tab:more_result_for_tab1}
        \scalebox{0.50}{
        \setlength{\tabcolsep}{2.5mm}{\begin{tabular}{c c c c} 
            \toprule[0.1em]
            Method & STQ & SQ & AQ \\
            \midrule[0.1em]
            Video K-Net &  33.1 &  35.0 & 29.6 \\
          \rowcolor{gray!15} Video K-Net + tube matching (Ours)  & 34.7 & 36.8 & 30.8 \\
           Video K-Net + tube matching (IFC) & 33.3 & 35.7 & 28.4 \\
            \bottomrule[0.1em]
        \end{tabular}}}
    \end{subtable}%
    \begin{subtable}{.5\linewidth}
        \centering
        \footnotesize
        \caption{KITTI-STEP test set results.}
        \label{tab:test_dev_kitti_step}
        \scalebox{0.80}{
        \setlength{\tabcolsep}{2.5mm}{\begin{tabular}{c c c } 
            \toprule[0.1em]
            Method & Backbone & STQ  \\
            \midrule[0.1em]
            Motion-Deeplab & ResNe50 & 0.52 \\
            Video K-Net & ResNet50 & 0.59 \\
             Video K-Net & ResNet50 & 0.63 \\
             \hline
             Tube-Link & ResNet50 & 0.60 \\
             Tube-Link & Swin-base & 0.65 \\
            \bottomrule[0.1em]
        \end{tabular}}}
    \end{subtable}
\end{table}

\subsection{More Ablations and Experiment Results}
\label{sec:more_ablation_and_exp_results}

In this section, we first present more detailed ablations for Tube-Link. Then, we present more detailed results on several datasets, including VIS datasets~\cite{vis_dataset}, OVIS dataset~\cite{qi2022occluded} and VSPW test set~\cite{miao2021vspw}.

\vspace{2mm}
\noindent
\textbf{More Ablations on Effectiveness of Tube-Wised Matching.}
In Tab.~\ref{tab:tube_wisedmethod}, we present more detailed ablations on tube size in Youtube-VIS. Note that, for simplicity, the input subclip size is the same as the tube size. As we enlarge the tube size, we find a significant improvement in the final performance. After enlarging the size to 4, the performance is the best. Using a tube size of 6, the performance slightly degrades. However, it still performs better than single-frame matching. All the models are trained under the same tube size (default is 2). The findings also verify our motivation for using clip-level matching, which shares similar findings on the VIP-Seg dataset in the main paper.

\vspace{2mm}
\noindent
\textbf{Inference with Overlapped Frames in VIP-Seg.} In Tab.~\ref{tab:over_lap_window_size}, we explore the effect of the overlapping size for nearby windows. As shown in that table, increasing the overlapping size leads to better performance for all three metrics: VPQ, STQ, and SQ. This is because we can use multiple frames twice, which leads to more consistent segmentation results. Moreover, instances in smaller windows are easier to be tracked. However, to save computation costs and increase inference speed, we do not introduce overlapping during inference. All the results in the main paper use non-overlapping inference.

\vspace{2mm}
\noindent
\textbf{Effect of COCO Pretraining in VIP-Seg.} In Tab.~\ref{tab:abl_coco_pretrain}, we show the effect of COCO pretraining on both Video K-Net+ and our Tube-Link. From the table, we can see that COCO pretraining plays an important role for VIP-Seg datasets, which shares the same conclusion with previous work~\cite{li2022videoknet,miao2022large}. Without COCO pretraining, both Video K-Net+ and Tube-Link drop a lot. However, as shown in the gray area, our method \textit{without} COCO pretraining outperforms the Video K-Net+ baseline by a large margin, where we still achieve over 8\% STQ gain and 14\% VPQ gain. The results suggest the effectiveness of our framework on better usage of temporal information.

\vspace{2mm}
\noindent
\textbf{Effect of Training Epoch on VIP-Seg.} We perform ablation on training epochs as in Tab.~\ref{tab:ablation_train_epoch}. With more training epochs, we do not observe performance gain with the COCO pre-trained model due to the overfitting issues. We use eight training epochs by default for all models.

\begin{figure*}[t!]
	\centering
	\includegraphics[width=0.85\linewidth]{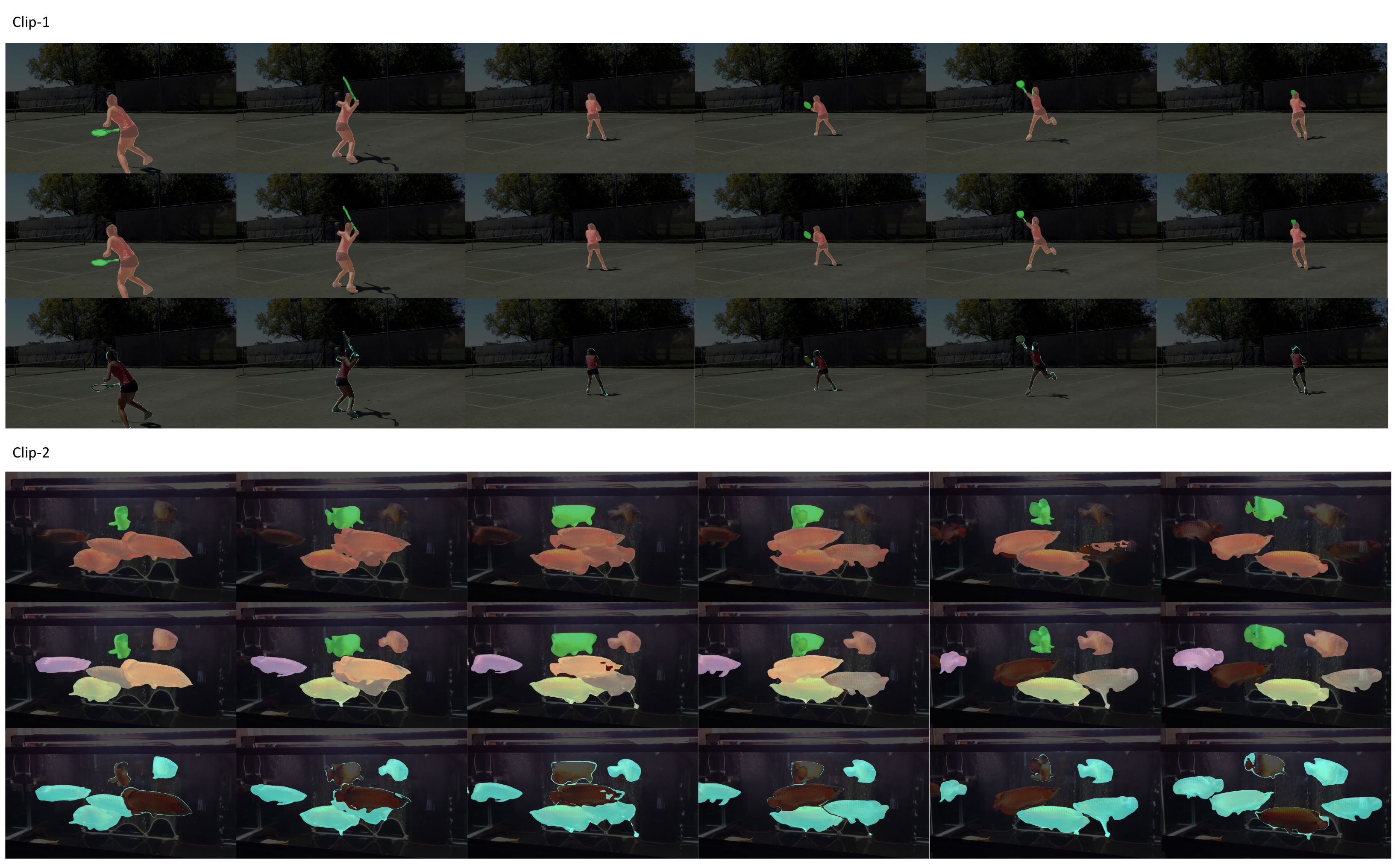}
	\caption{Visual Comparison Results from Tube-Link with ResNet50 backbone. Our method (middle) achieves consistent segmentation and better segmentation/tracking results than the Mask2Former-VIS baseline (top). We also visualize the difference maps (bottom). \textbf{Best viewed by zooming in.}}
	\label{fig:yt_vis_2019_comparison}
\end{figure*}

\begin{figure*}[t!]
	\centering
	\includegraphics[width=0.85\linewidth]{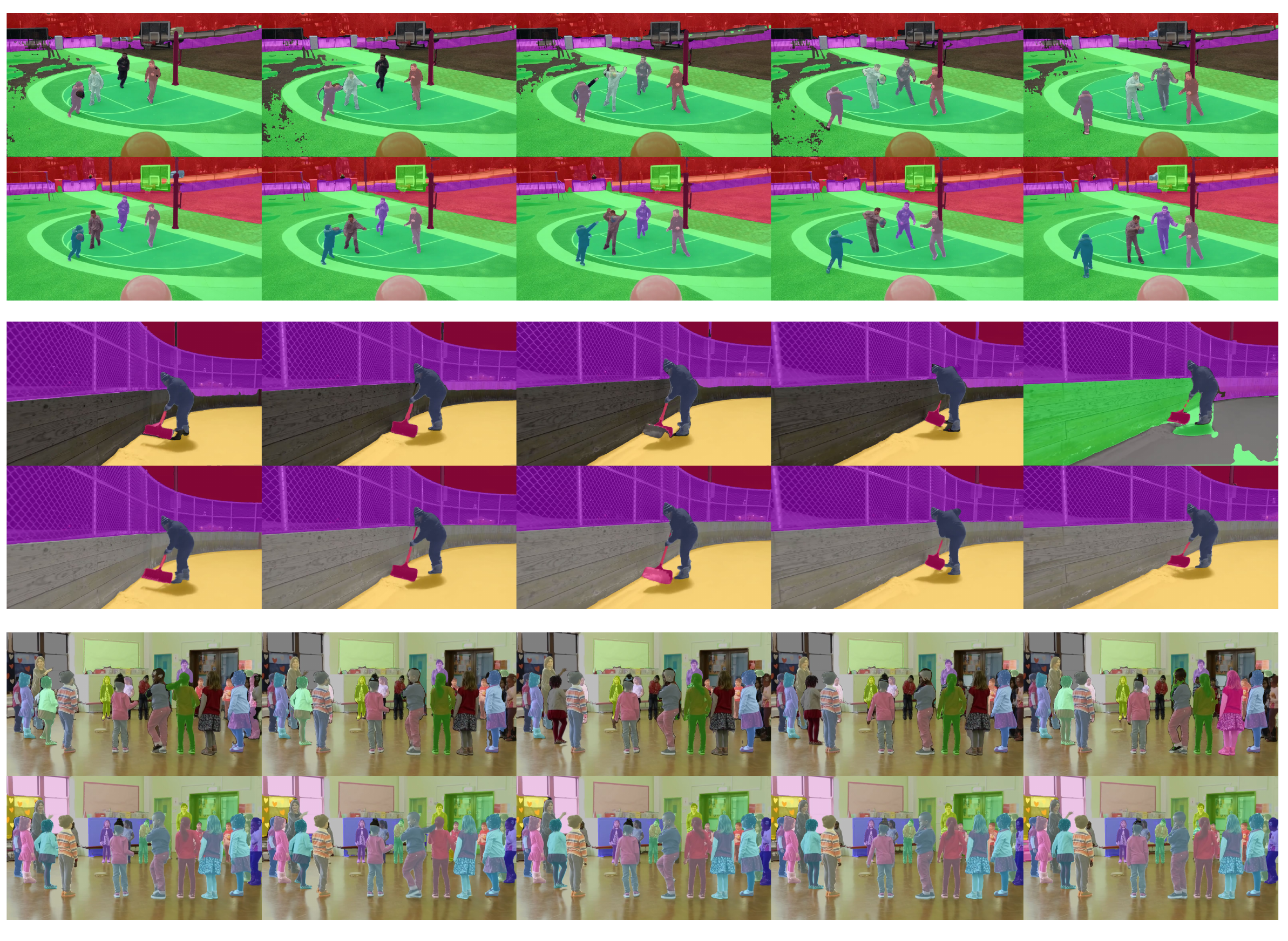}
	\caption{More Visual Results from Tube-Link with ResNet50 backbone. Our method (top) achieves consistent segmentation and better tracking results than the Video K-Net+ baseline (bottom). \textbf{Best viewed by zooming in.}}
	\label{fig:more_vis_vip_seg}
\end{figure*}

\begin{figure*}[t!]
	\centering
	\includegraphics[width=0.85\linewidth]{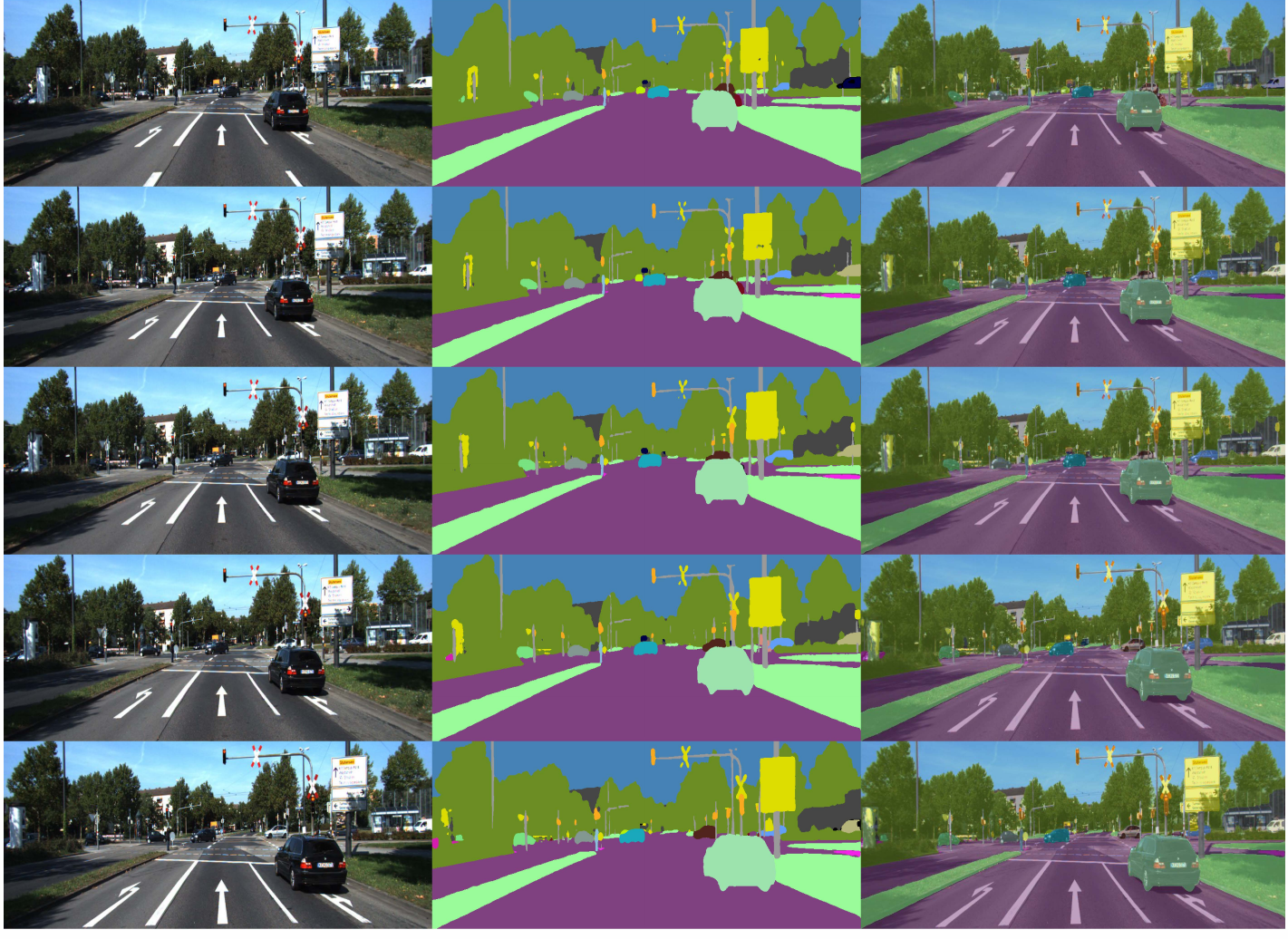}
	\caption{Visual Results from Tube-Link with ResNet50 backbone on the KITTI-STEP dataset. \textbf{Best viewed by zooming in.}}
	\label{fig:more_vis_kitti_step}
\end{figure*}

\begin{figure*}[t!]
	\centering
	\includegraphics[width=0.85\linewidth]{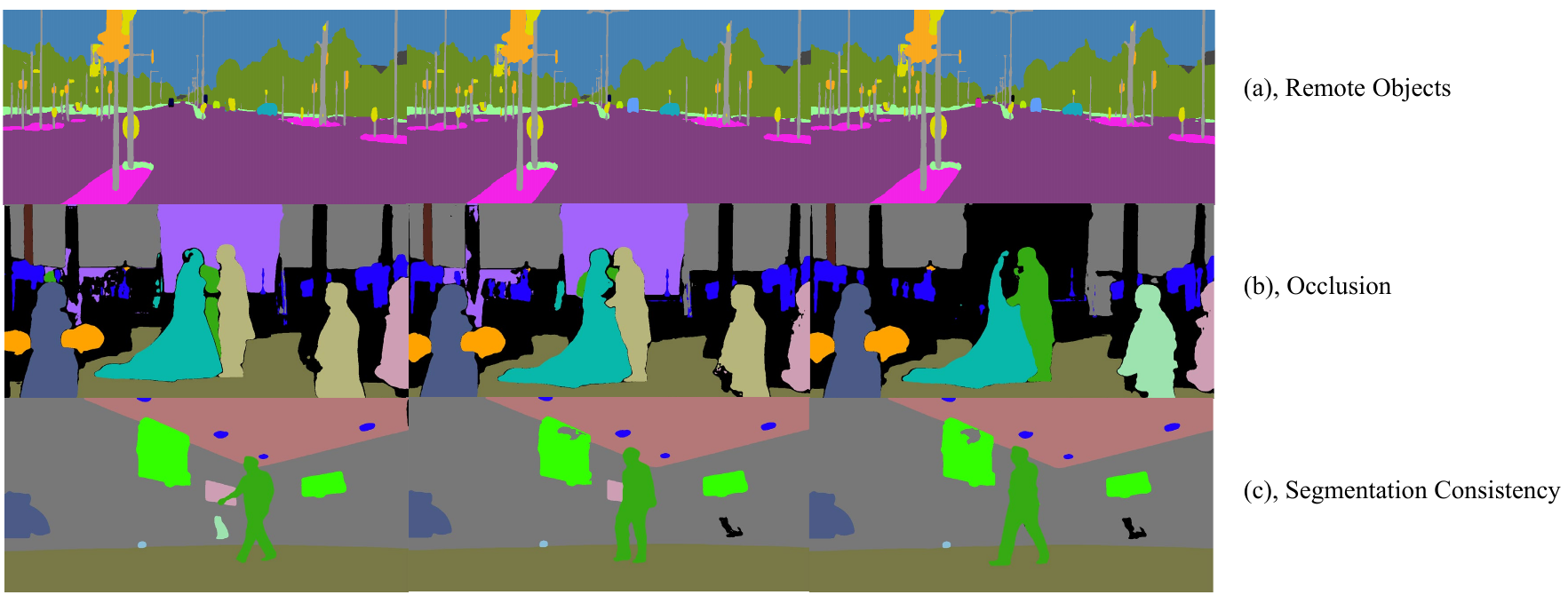}
	\caption{Visual Results on Failure Cases of Tube-Link. (a), Remote objects lead to ID switches and inferior segmentation results. (b), Heavy occlusion leads to an ID switch. (c), Segmentation consistency problems caused by camera motion.}
	\label{fig:failure_cases}
\end{figure*}

\begin{table}[!t]
        \centering
        \footnotesize
        \caption{Effect Of Quasi-Dense Tracker (Results on Youtube-VIS-2019 validation set).}
        \label{tab:effect_of_quasi_dense_tracker}
        \scalebox{0.80}{
        \setlength{\tabcolsep}{2.5mm}{\begin{tabular}{c c c c} 
            \toprule[0.1em]
            Method & Naive Tracker & Quani-Dense Tracker &  mAP \\
                  \midrule[0.1em]
            MinVIS  &  \checkmark & -  & 47.4 \\
            MinVIS  &  - & \checkmark  & 48.0 (+0.6) \\
            MiniVIS + tube matching &  \checkmark & - & 48.8 (+1.4) \\
            \hline
            Tube-Link &  \checkmark & - &  52.6 (-0.2) \\
            Tube-Link &  - & \checkmark & 52.8 \\
            \bottomrule[0.1em]
        \end{tabular}}}
\end{table}

\noindent
\textbf{Impact of Quasi-Dense Tracker.}
We adopt the same quasi-dense tracker for all experiments in the main paper, and we can achieve 3.0\% VPQ improvement upon the baseline. In Tab.~\ref{tab:effect_of_quasi_dense_tracker}, we perform an extra experiment by replacing our tracker with a naive tracker used in MinVIS, where we only found 0.2\% mAP drop. This proves the robustness and generalizability of Tube-Link. In contrast, we add the quasi-dense tracker to MinVIS, and we only find 0.6\% mAP improvements. Directly extending a method with tube matching leads to more improvements. The results also indicate that the effect of the tracker is not apparent on the Youtube-VIS dataset, since the instance number is limited and occlusion is not heavy. Thus, we adopt \textit{simple tube-matching} for VIS datasets. 

\vspace{2mm}
\noindent
\textbf{Detailed Results Youtube-VIS.} In Tab.~\ref{tab:ytvis_supp}, we report the detailed results on Youtube-VIS-2019 and Youtube-VIS-2021 datasets. We follow the baseline method, Mask2Former-VIS~\cite{cheng2021mask2former_vis}. 
As shown in that table, our method achieves all the best metrics on both datasets \textit{without COCO video joint training or clip-wised copy-paste}.

\noindent
\textbf{More Results on Test Set.}. Moreover, we also report our results on the KITTI-STEP test set. As shown in Tab.~\ref{tab:test_dev_kitti_step}, Our method can still achieve better results.

\vspace{2mm}
\noindent
\textbf{Detailed Results on OVIS.} In Tab.~\ref{tab:ovis}, we also report our model results on OVIS. Again, without bells and whistles, our method achieves comparable results with IDOL. We use the ResNet50 backbone for a fair comparison.

\subsection{Visual Results}
\label{sec:vis_results}

\noindent
\textbf{Visual Comparison on Youtube-VIS-2019 dataset.} In Fig.~\ref{fig:yt_vis_2019_comparison}, we compare our Tube-Link with strong baseline Mask2Former-VIS with the same ResNet50 backbone. Our methods achieve more consistent tracking and segmentation results in two examples.

\noindent
\textbf{More Visual Results on VIP-Seg Dataset.} In Fig.~\ref{fig:more_vis_vip_seg}, we present more visual examples on our Tube-Link. Compared with the Video K-Net+ baseline, our method achieves better segmentation and tracking consistency.

\noindent
\textbf{Visual Results on KITTI-STEP Dataset.} In Fig.~\ref{fig:more_vis_kitti_step}, we present visual results on the KITTI-STEP dataset, where we achieve consistent segmentation and tracking on the driving scene.

\noindent
\textbf{Failure Cases Analysis.} In Fig.~\ref{fig:failure_cases}, we show several failure cases on the KITTI-STEP and VIP-Seg datasets using our best models. We observe three error sources: (1). remote and small objects. (2). heavy occlusion. (3). segmentation consistency caused by camera motion. We will handle these issues in future work.

{\small
\bibliographystyle{ieee_fullname}
\bibliography{egbib}
}

\end{document}